\documentclass[aps,pre,floatfix,epsf,epsfig,nofootinbib]{revtex4} 
\usepackage{dcolumn}
\usepackage{slashbox}
\usepackage{bm}
\usepackage[dvips]{epsfig}
\usepackage{graphicx} \usepackage{amsmath} \usepackage{amssymb}
\usepackage{epsfig}
\newcommand{\BEQ}{\begin{equation}}
\newcommand{\EEQ}{\end{equation}}
\newcommand{\BEA}{\begin{eqnarray}}
\newcommand{\EEA}{\end{eqnarray}}

\renewcommand{\d}{{\rm d}}

\newcommand{\comment}[1]{}

\begin{document} 
\draft

\title{Observational nonidentifiability, generalized likelihood and free energy}

\author{A.E. Allahverdyan}
\affiliation{ Yerevan Physics Institute, Alikhanian Brothers Street 2, Yerevan 375036, Armenia}

\begin{abstract} We study the parameter estimation problem in mixture
models with observational nonidentifiability: the full model (also
containing hidden variables) is identifiable, but the marginal
(observed) model is not. Hence global maxima of the marginal likelihood
are (infinitely) degenerate and predictions of the marginal likelihood
are not unique. We show how to generalize the marginal likelihood by
introducing an effective temperature, and making it similar to the free
energy. This generalization resolves the observational
nonidentifiability, since its maximization leads to unique results that
are better than a random selection of one degenerate maximum of the
marginal likelihood or the averaging over many such maxima.  The
generalized likelihood inherits many features from the usual likelihood,
e.g. it holds the conditionality principle, and its local maximum can be
searched for via suitably modified expectation-maximization method. The
maximization of the generalized likelihood relates to entropy
optimization.


\end{abstract}

\maketitle


\section{Introduction}

Unknown parameters of mixture models are frequently estimated via the
Maximum Marginal Likelihood (MML) method that employs the marginal
probability of the observed data
\cite{pawitan,cox,jelinek_review,rabiner_review,ephraim_review}. A local
maximization of the marginal likelihood can be carried out via one of
computationally feasible algorithms, e.g. the Expectation-Maximization
(EM) method \cite{jelinek_review,rabiner_review,ephraim_review}. 

There is however a range of problems, where MML does not apply due to
{\it observational nonidentifiability}: the full model (including hidden
variables) is identifiable, but the observed (marginal) model is not.
Hence the maxima of the marginal likelihood are (generally infinitely)
degenerate, and the outcome of MML does depend on the initial point of
the maximization. Resolving the nonidentifiability in such situations is
not hopeless, precisely because the full model is identifiable. However,
the standard likelihood maximization cannot be employed, since there are
hidden (not observed) variables. We emphasize that some information about
unknown parameters is always lost after marginalization \cite{cox}. The
observational nonidentifiability is an extreme case of this. 

Nonidentifiability in mixture models is studied in
\cite{teicher,rothenberg,ito,watanabe,hsiao,ran_hu,welcher,manski,allman,gu}; see
\cite{hsiao,ran_hu,welcher} for reviews. In such models
even an infinitely large number of observed data samples cannot
guarantee the perfect recovery of parameters (i.e. the convergence to
true parameter values), because maxima of the likelihood are infinitely
degenerate \cite{rothenberg}. There is an attitude towards
nonidentifiable models that they are in a certain sense rare, and do
not have a big practical importance. This is incorrect: almost any model
becomes nonidentifiable if the number of unknown parameters is
sufficiently large, i.e. if the model is sufficiently realistic
\cite{watanabe,hsiao}. Moreover, nonidentifiability can be present
effectively due to unresponsiveness of a many-parameter likelihood along
sufficiently many directions \cite{sethna1,sethna2}; see \cite{sethna3} for a review. The
simplest scenario of this is realized via small eigenvalues of the
likelihood Hessian. For practical purposes such an effective
nonidentifiability|which is generically found in systems biology and chemistry
\cite{sethna1,sethna2,sethna3}|is indistinguishable from the true one. 

\comment{
One way of dealing with the observational nonidentifiability is to
employ the full likelihood maximization both over the unknown variables
and over the hidden (not observed) variables
\cite{pawitan,jelinek_review,rabiner_review,ephraim_review}. In
statistics this method is known as the h-likelihood
\cite{nelder,bjorn,scand}, while in the field of Hidden Markov Models
(HMM) people call it Viterbi Training (VT) or k-means segmentation
\cite{jelinek_review,rabiner_review,ephraim_review,rabiner,merhav}. It
is also employed without observational nonidentifiability, since it
provides advantages in concrete applications
\cite{jelinek_review,rabiner_review,ephraim_review}. In fact, this
method employs the full probability of the mixture model, where
(unknown) hidden variables are replaced by their maximum-aposteriori
(MAP) estimates \cite{nelder,bjorn,scand}. VT can also be implemented
locally similar to the EM algortihm \cite{byrne}. 
}

Aiming to solve the problem of observational nonidentifiability, we
extend the marginal likelihood via a one-parameter generalized function
${L}_\beta$, which is constructed by analogy to the free energy in
statistical physics. The positive parameter $\beta$ is an analogue of
the inverse temperature from statistical physics, and the marginal
likelihood is recovered for $\beta=1$.  We show that ${L}_\beta$
inherits pertinent features of ${L}_1$; e.g. it holds the conditionality
principle, concavity (for $\beta\leq 1$) and the possibility to search
for its local maximum via suitably generalized expectation-maximization
method.  Its maximization resolves the degeneracy of $L_1$.  It does
have relations with the maximum entropy method (for $\beta< 1$) and with
entropy minimization (for $\beta>1$). For several models we found an
optimal value of $\beta$ in ${L}_\beta$, which appears to be close to 1,
but strictly smaller than $1$.  We also show numerically that maximizing
${L}_{\beta\lesssim 1}$ leads to better results than {\it (i)} a random
selection of one of many results provided by maximizing the usual
likelihood ${L}_1$; {\it (ii)} averaging over many such random
selections; see section \ref{compo}. Both {\it (i)} and {\it (ii)} would
be among standard reactions of practitioners to (effective)
nonidentifiability. 

For $\beta\to\infty$ we get another known quantity: ${L}_\infty$
coincides with the h-likelihood
\cite{pawitan,jelinek_review,rabiner_review,ephraim_review}, i.e.  the
full likelihood (including both observed and hidden variables), where
the value of hidden variables is replaced by their maximum aposteriori
(MAP) estimates from the observed data \cite{nelder,bjorn,scand}. The
h-likelihood is employed in Hidden Markov Models (HMM), where efficient
methods of maximizing ${L}_\infty$ are known as Viterbi Training (VT) or
k-means segmentation
\cite{jelinek_review,rabiner_review,ephraim_review,rabiner,merhav}.
When the h-likelihood ${L}_\infty$ is applied to an observationally
nonidentifiable situation, its results converge to boundary values of
the parameters (e.g. zero or one for unknown probabilities), as was
demonstrated by analyzing an exactly solvable HMM model \cite{nips}.
Such results are inferior to random selection (see {\it (i)} and {\it
(ii)} above), if there is no prior information that the model is indeed
sparse in this sense; cf.~section \ref{beta>1}. This feature is
one reason why the h-likelihood maximization leads to obvious failures
even in simple models \cite{nelder,meng}. In particular, it cannot apply
generally for solving observational nonidentifiability.

\comment{ Another common prejudice against the nonidentifiability
phenomenon is that once a certain function of unknown parameters can be
determined uniquely, the situation is not different from the
identifiable situation. This opinion is incorrect as well: even with
respect to a uniquely defined parameter, point estimates in
nonidentifiable models do differ from identifiable ones
\cite{watanabe}. 

Once for nonidentifiable models no perfect recovery of
parameters is possible (even in the limit of a large number of data
samples), standard learning methods can give only certain bounds for
unknown parameters. In practice, people employ these bounds
supplementing them with various ad hoc methods. Here we should mention
another (third) incorrect opinion about nonidentifiability: people
believe that all outcomes of the maximum-likelihood method for a
nonidentifiable model are equivalent; hence finding sufficiently rigid
bounds on them-e.g. via confidence intervals-can be sufficient for their
characterization. This is incorrect as exemplified by the
over-confidence phenomenon: one serious obstacle in applying the outcomes
of the maximum-likelihood method for a nonidentifiable model is that
they can violate most obvious conditions on the noise-filtering. For
example, they can lead to Bayesian risks lower than their minimal
theoretical value, thereby creating an over-confidence on the
performance of filtering [12]. }

$L_\beta$ also relates to a recent trend in the Bayesian statistics,
where the model is raised to a certain positive power, akin to
$\beta\not=1$ in $L_\beta$ \cite{bissiri,holmes,hagan,friel,miller}.  In
this way people deal with misspecified models
\cite{bissiri,holmes,miller}, facilitate the computation of Bayesian
factors for model selection \cite{friel}, regularize them \cite{hagan}
{\it etc}; see \cite{miller} for a recent review. The raising into a
power emerges from the decision theory (as applied to misspecified
models) \cite{bissiri} and present a general method for making Bayesian
models more robust. Among the actively researched issues here is the
selection of the power parameter \cite{holmes}. 

This paper is written in the style of the book by Cox and Hinkley
\cite{cox}: it is example-based and informal, not the least because it
employs ideas of statistical physics. It is organized as follows.
Section \ref{defo} recalls the definition of the observational
nonidentifiability we set to study.  Section \ref{gena} defines the
generalized likelihood $L_\beta$, and discusses its features inherited from
the usual likelihood $L_1$. Sections \ref{simpo} and \ref{counter} study
the simplest nonidentifiable examples that illustrates features of
$L_\beta$. Section \ref{mixo} defines the main model we shall focus on.
It amounts to a finite mixture with unknown probabilities.  Section
\ref{beta<1} studies for this model the generalized likelihood
$L_{\beta<1}$. Numerical comparison with the random selection methods is
discussed in section \ref{compo}. Section \ref{beta<1} studies the
maximization of $L_{\beta>1}$ and shows in which sense this is related
to entropy minimization. We summarize in the last section. 

\section{Free energy as generalized likelihood}
\label{free_energy}

\subsection{ Defining observational nonidentifiability }
\label{defo}

We are given two random variables $X$ and $Y$ with values $x=(x_1,...,x_n)$
and $y=(y_1,...,y_m)$, respectively. We assume that $X$ is hidden, while $Y$
observed variable, i. e. we assume a mixture model. The joint probabilities of $XY$
\BEA
\label{1}
p_\theta(x,y),
\EEA
generally depend on unknown parameters $\theta$.
Let we are given the observation data
\BEA
\label{data}
\{ \,y^{[k]}\,\}_{k=1}^N,
\EEA
where $y^{[k]}$ are values of $Y$ generated independently from each other. Then $\theta$ can be estimated via 
the (marginal, logarithmic) likelihood 
\BEA
\label{44}
L(\theta)=\frac{1}{N}\sum_{k=1}^N \ln\left[p_\theta(y^{[k]}) \right]=\sum_yp(y)\ln\left[\sum_x p_\theta(x,y) \right],
\label{4}
\EEA
where $ p(y_1),...,  p(y_m)$ are the frequencies of $Y$ obtained from the data (\ref{data}). 
If (for a fixed $m$) the observation data (\ref{data}) is large, $N\gg 1$,
$p(y)$ converge to true probabilities of $Y$. 

Within the maximum likelihood method, the unknown $\theta$ can be
determined from ${\rm argmax}_\theta [\,L(\theta)]$.  Since $X$ is a
hidden, we can easily run into the nonidenitifiability problem, where
(at least two) different values of $\theta$ lead to the same probability
for all values of $Y$ \cite{teicher,rothenberg,ito,hsiao}:
\BEA
 p_\theta(y_l)= p_{\theta'}(y_l), \quad l=1,...,m, \quad \theta\not=\theta'.
\label{nono}
\EEA
Eqs.~(\ref{nono}) imply that maxima of $L(\theta)$ are degenerate; see
below for examples. In addition to (\ref{nono}), we shall require that
the full model is still identifiable, i.e. imposing equal joint
probabilities for for all $(x,y)$ does lead to $\theta=\theta'$ for all
$(\theta , \theta ')$:
\BEA
\label{star}
p_\theta(x_k,y_l)= p_{\theta'}(x_k,y_l), \quad k=1,...,n, ~  l=1,...,m,\quad \Longrightarrow\quad \theta=\theta'.
\EEA
We shall propose a solution to this type of nonidentifiability. Below we
shall focus the most acute situation, where the marginal probability in
(\ref{nono}) does not depend on $\theta$ and the sample length in (\ref{data}) 
is very large: $N\gg 1$. Note that other (weaker) forms
of nonidentifiability are possible and well-documented in literature:
the weakest form of nonidentifiability is when it is restricted to a
measure-zero subset of the parameter domain (generic identifiability) 
\cite{allman}. A stronger
form is that of partial nonidentifiability, where some information
on $\theta$ (e.g. certain bounds on $\theta$) can still be recovered 
from observations; see \cite{gu} for a recent discussion. 

\comment{ Thus we need selection criteria for how to choose the simplest
and/or the most reasonable one among those maxima. And we need
meta-selection criteria for how to define what is the simplest and the
most reasonable. }

\subsection{Generalized likelihood: definition and features}
\label{gena}

\subsubsection{Definition }

Instead of (\ref{4}) we set to maximize over $\theta$
its generalization, {\it viz.} the negative free-energy
\BEA
\label{6}
L_\beta(\theta)=\frac{1}{\beta}\sum_yp(y)\ln\left[\sum_x p_\theta^\beta(x,y) \right],
\EEA
where $\beta>0$ is a parameter. 
An obvious feature of (\ref{6}) is that for $\beta=1$ we return from
(\ref{6}) to the (marginal) likelihood function $L(\theta)$ in
(\ref{4}). Hence if we apply the maximization of (\ref{6}) with
$\beta\approx 1$ to the identifiable model, we expect to get results
that are close to those found via maximization of $L(\theta)$. The
meaning of $\beta\not=1$ in (\ref{6}) is that it sums over all values of
$x$, but does not reduce the outcome to the usual (marginal) likelihood. 

Below we discuss several features that $L_\beta(\theta)$
inherits from the usual likelihood $L_1(\theta)=L(\theta)$. These
features motivate introducing $L_\beta(\theta)$ as a generalization of
$L_1(\theta)$. The first such feature is apparent from 
the fact that $L_\beta(\theta)$ in (\ref{6}) is to be maximized over 
unknown parameter $\theta$
\cite{pawitan}. If we reparametrize $L_\beta$ via a bijective (one-to-one)
function $\psi(\theta)$|i.e. if the full information on $\theta$ is retained 
in $\psi$|then the maximization outcomes
\BEA
\hat\psi={\rm argmax}_\psi [L_\beta(\psi)] \quad {\rm and}\quad
\hat\theta={\rm argmax}_\theta [L_\beta(\theta)], 
\EEA
are related via the same function: $ \hat\psi=\psi(\hat \theta)$.

\comment{Whenever the dependence on $XY$ is to be made explicit, we write
$L^{[XY]}_\beta(\theta)$ instead of $L_\beta(\theta)$.}

\subsubsection{Relations of (\ref{6}) to nonequilibrium free energy}

Relations between statistical physics and probabilistic inference
frequently proceed via the Gibbs distribution, where the minus logarithm
of probability to be inferred is interpreted as the physical energy
(both these quantities are additive for independent events), while the
physical temperature is taken to be $1$; see \cite{mezard} for a
textbook presentation of this analogy and \cite{lamont} for a recent
review.  The main point of making this analogy is that powerful
approximate methods of statistical physics can be applied to inference
\cite{mezard,lamont}. 

In the context of mixture models we can carry out the above analogy one step
further. This analogy is now structural, i.e. it relates to the form of
(\ref{6}), and not to applicability of any approximate method. We relate
$-\ln\hat p(x,y)$ with the energy of a physical system, where $X$ and
$Y$ are respectively fast (hidden) and slow (observed) variables. Here
fast and slow connect with (resp.) hidden and observed, which agrees
with the set-up of statistical physics, where only a part of variables
is observed \cite{free}. Then (\ref{6}) connects to the negative
nonequilibrium free energy with inverse temperature $\beta$ \cite{free}.  Here
nonequilibrium means that only one variable (i.e. $X$) is thermalized 
(i.e. its conditional probability is Gibbsian), while
the free energy has several physical meanings \cite{free}; e.g. it is a generating
function for calculating various averages and also the (physical) work
done under a slow change of suitable externally-driven parameters
\cite{free}. The maximization of (\ref{6}) naturally relates to the
physical tendency of decreasing free energy (one formulation of the
second law of thermodynamics) \cite{free}. 

Though formal, this correspondence with statistical physics will be
instrumental in interpreting $L_\beta$. E.g. we shall see that the
maximizer of $L_{\beta<1}$ is unique (in contrast to maximizers of
$L_{\beta\geq 1}$), and this fact can be related to sufficiently high
temperatures that simplify the free energy landscape.

\subsubsection{Relations with h-likelihood}

For $\beta\to\infty$ we revert from (\ref{6}) to
\BEA
L_\infty(\theta)=\sum_yp(y)\ln\left[{\rm max}_x \, p_\theta(x,y)\right],
\label{udodo}
\EEA
where ${\rm argmax}_x\hat p_\theta(x,y)$ is the MAP (maximum
aposteriori) estimate of $x$ given the data $y$
\cite{ephraim_review,rabiner,nips}. The meaning of (\ref{udodo}) is
obvious in the context of (\ref{star}): once we cannot employ the
maximum likelihood method to $p_\theta(x,y)$|since we do not know what
to take for the hidden variable $x$|we first estimate $x$ from data
(\ref{data}) via the MAP method, and then proceed {\it a la} usual
likelihood \footnote{Note that in (\ref{udodo}) the maximization was
carried out for a given value of $y$, i.e. we did not apply it to the
whole sample (\ref{data}). Doing so will lead to $L'_\infty(\theta)={\rm
max}_x \left[\,\sum_yp(y)\ln p_\theta(x,y)\right]$ instead of
(\ref{udodo}). We did not see applications of $L'_\infty(\theta)$ in
literature.  One possible reason for this is that the definition of
$L'_\infty(\theta)$ makes an unwarranted (though not strictly forbidden)
assumption that $X=x$ is fixed during the sample generation process. At
any rate, we applied $L'_\infty(\theta)$ to models and noted that its
results for parameter estimation are worse than those of
$L_\infty(\theta)$.  Hence we stick to (\ref{udodo}).}. 

It is known that maximizing over unobserved variables has drawbacks \cite{pawitan,nelder,meng}. 
People tried to improve on those drawbacks by looking instead of (\ref{udodo}) at
\cite{byrne}
\BEA
\frac{1}{U}\sum_{u=1}^U\sum_y p(y)\ln\left[ p_\theta (x_{[u]}(y),y)
\right],
\label{faso}
\EEA 
where $U=2,3$, $x_{[1]}(y)$ maximizes $p(x,y)$ over $x$, $x_{[2]}(y)$ is the next to
the maximal value of $p(x,y)$ {\it etc}. In contrast to (\ref{udodo}),
Eq.~(\ref{faso}) accounts for values of $x$ around the maximum of
$p(x,y)$. Now $L_\beta(\theta)$ captures the same idea for a large but finite
$\beta$.

\subsubsection{Conditionality}

It is known that the ordinary maximum-likelihood method has an appealing
feature of conditionality, which is formulated in several related forms
\cite{cox}, and closely connects to other fundamental principles of
statistics, e.g. to the likelihood principle \cite{cox,berger,evans}. We now find
out to which extent the conditionality principle is inherited by the
generalized likelihood $L_\beta$ defined in (\ref{6}).

First we note that $L_\beta$ holds the weak conditionality
principle \cite{berger}. To define this principle we should enlarge the
original pair $(X,Y)$ of random variables to $(X,Y,J)$, where $J$
assumes (for simplicity) a finite set of values $j_1,..,j_\ell$. Now $X$
and $Y$ are still (resp.) hidden and observed variables, while $J$
determines the choice of the experiment that does not depend on the
unknown parameter $\theta$ \cite{berger,evans}. The choice is done before
observing $Y$, i.e. before collecting the sample (\ref{data}), and the
(marginal) probability $p(j)$ does not depend on $\theta$. 
For this extended experiment the data amounts to sample (\ref{data}) plus 
the indicator $j$ for the choice of the experiment. Then the
analogue of (\ref{6}) is defined as
\BEA
{L}_\beta(j,\theta)=\frac{1}{\beta}\sum_yp(y)\ln\left[\sum_x p^\beta_\theta(x,y,j)\right]
=\frac{1}{\beta}\sum_yp(y)\ln\left[\sum_x p^\beta_\theta(x,y|j)\right]+\ln p(j),
\label{ushi}
\EEA
where $p_\theta(x,y,j)$ is the probability of $(X,Y,J)$. 
It is seen that the inference for the extended
experiment produces the same result as the inference 
for the partial experiment, where the value of $J=j$ was fixed beforehands (i.e. the choice of $J=j$
was not a part of data): 
\BEA
{\rm argmax}_\theta\left({L}_\beta(j,\theta)\right)={\rm argmax}_\theta\left(
\frac{1}{\beta}
\sum_yp(y)\ln\left[\sum_x p^\beta_\theta(x,y|j)\right]\right).
\EEA
This is the weak conditionality principle that holds for the generalized likelihood $L_\beta(\theta)$. 

However, a stronger form of the conditionality principle does not hold
for $L_\beta(\theta)$, because this form mixes observable and hidden variables.
Define a new random variable $G(X,Y)$ that depends on $X$ and $Y$ and
assumes values $g_1,...,g_\ell$ \cite{cox}. Assume that the marginal
probability $p(g)$ of $G$ does not depend on $\theta$, i.e. $G$ is an
ancillary variable with respect to estimating $\theta$ 
\cite{fraser_review} \footnote{Recall that ancillary variables 
need not always exist for a given model \cite{sze}.}. Now (\ref{6}) reads
\BEA
{L}_\beta(\theta)=\frac{1}{\beta}
\sum_yp(y)\ln\left[\sum_{g\in {\cal G}(y)} p^\beta(g)\sum_{x:\, G(x,y)=g} p^\beta_\theta(x,y|g)\right],
\label{ushov}
\EEA
where ${\cal G}(y)\subset (g_1,...,g_\ell)$ is the set of values assumed by
$G(x,y)$, where $y$ is fixed, while $x$ goes over all its values
$(x_1,..,x_n)$.  One defines a new experiment, where it is {\it a
priori} known that the value of $G(X,Y)$ is restricted to a specific value $g$ from
$(g_1,...,g_\ell)$. The generalized likelihood for this experiment
is
\BEA
{L}_\beta(\theta|G=g)=\frac{1}{\beta}
\sum_yp(y)\ln\left[\sum_{x:\, G(x,y)=g} p^\beta_\theta(x,y|g)\right].
\label{usho}
\EEA
It is seen that for $\beta\not=1$ the maximization of (\ref{ushov}) and (\ref{usho})
will generally produce different results, i.e. the
stronger form of the conditionality principle does not hold for $L_\beta$. 

\comment{
$L_\beta(\theta)$ is additive for independent random variables $(X,Y)$ and $(\tilde{X},\tilde{Y})$, i.e.
\BEA
p_\theta(x,\tilde{x},y,\tilde{y})= p_\theta(x,y) p_\theta(\tilde{x},\tilde{y}), \qquad p(y,\tilde{y})=p(y)p(\tilde{y}),
\EEA
leads from (\ref{6}) to
\BEA
\label{66}
L^{[XY\tilde{X}\tilde{Y}]}_\beta(\theta)=L^{[XY]}_\beta(\theta)+L^{[\tilde{X}\tilde{Y}]}_\beta(\theta).
\EEA
Eq.~(\ref{66}) has implications related to the conditionality
principle \cite{pawitan,cox,berger}: if $p_\theta(\tilde{x},\tilde{y})$
(and hence $L^{[\tilde{X}\tilde{Y}]}_\beta$) does not depend on
$\theta$, then $L^{[\tilde{X}\tilde{Y}]}_\beta$ enters into (\ref{66})
as an additive factor that drops out when looking at differences of
$L^{[XY\tilde{X}\tilde{Y}]}_\beta(\theta)$ as a function of $\theta$.
Thus only variables whose probability depends on $\theta$ can contribute
into estimation of $\theta$ via $L_\beta(\theta)$. The conditionality
principle implies a well-known fact that changes of $L_\beta(\theta)$
depend only on a sufficient statistics of data \cite{cox}.
}

\subsubsection{Monotonicity and concavity}

$L_\beta (\theta)$ is monotonically decreasing over $\beta$:
\BEA
&& \frac{\partial L_{\beta}(\theta)}{\partial \beta}=\frac{1}{\beta^2}\sum_y p(y)\sum_x \zeta_\theta(x|y;\beta)\ln \zeta_\theta(x|y;\beta)\leq 0,\\
&& \zeta_\theta(x|y;\beta)\equiv \left. { p_\theta^\beta(x,y)}\right/{{\sum}_{\bar{x}}  p_\theta^\beta(\bar x, y)  },
\label{drud}
\EEA
since $\frac{\partial L_{\beta}(\theta)}{\partial \beta}$ is a weighted
sum of negative entropies. 

Let $\theta\in\Omega$ is defined over a partially convex set $\Omega$,
i.e. if $\theta_1\in\Omega$ and $\theta_2\in\Omega$, then for $0<
\lambda< 1$ there exists $\theta_3\in\Omega$ such that
$p_{\theta_3}=\lambda p_{\theta_1}+(1-\lambda) p_{\theta_2}$; such a
model is studied below in section \ref{mixo}. Now for $\beta\leq 1$,
$L_\beta$ from (\ref{6}) is a concave function, since it is a linear
combination of superposition of two strictly concave functions:
$f(u)=u^\beta$ and $g(v)=\ln v$:
\BEA
L_{\beta}(\lambda p_{\theta_1}+(1-\lambda) p_{\theta_2})>\lambda L_{\beta}(p_{\theta_1})+(1-\lambda) L_{\beta}(p_{\theta_2}), \qquad \beta\leq 1.
\label{glinka}
\EEA

For $\beta>1$, we note that a superposition $g(f(u))$ of strictly convex
$f(u)=u^\beta$ and monotonic $g(v)=\ln v$ is pseudo-convex
\cite{pseudoconvex}. Pseudo-convex functions do share many important
features of convex functions, but generally $L_{\beta>1}$ is not
pseudo-convex, since besides superposition of $f(u)=u^\beta$ and
$g(v)=\ln v$, (\ref{6}) involves summation over $y$, and the sum of two
pseudo-convex functions is generally {\it not} pseudo-convex
\cite{pseudoconvex}. In section \ref{beta>1} we shall show numerically
that maximizers of $L_{\beta>1}$ relate to those of a generalized
Schur-convex function; see Appendix \ref{schuro}. 

\comment{
\BEA
L_{\beta>1}(\lambda\hat p_1+(1-\lambda)\hat p_2)=
\sum_y p(y)\ln\sum_x[\lambda\hat p_1(x,y)+(1-\lambda)\hat p_2(x,y)]^\beta\\
<\sum_y p(y)\,
{\rm max}\left[  \ln\sum_x\hat p^\beta_1(x,y),\,\, \ln\sum_x\hat p^\beta_2(x,y)  
\right] = {\rm max}\left[  \ln\sum_x\hat p^\beta_1(x,y),\,\, \ln\sum_x\hat p^\beta_2(x,y)  
\right]
\EEA }

\comment{
The fact of concavity is important: as known from the convex analysis,
if we maximize $L_\beta$ over a convex domain, and find a stationary
point (i.e. where the first derivatives of $L_\beta$ vanish) that is an
internal point (i.e. it is not on the vertices of the convex domain),
then this stationary point is a global maximum of $L_\beta$. Given these useful 
features, in the present section we restricted ourselves to $0<\beta\leq 1$.}

\subsubsection{Relations with the maximum entropy method}
\label{uraa}

The maximization of the generalized likelihood (\ref{6}) will be now
related with the maximum entropy method
\cite{jaynes,jaynes_2,skyrms,enk,cheeseman}.  Recall that the method
addresses the problem of recovering unknown probabilities
$\{q(z_k)\}_{k=1}^n$ of a random variable $Z=(z_1,...,z_n)$ on the
ground of certain contraints on $q$ and $Z$. The type and number of
those constrains are not decided within the method itself
\cite{jaynes_2,enk}, though the method can give some recommendations for
selecting relevant constraints; see Appendix \ref{tra}.  Then
$\{q(z_k)\}_{k=1}^n$ are determined from the constrained maximization of
the entropy $-\sum_{k=1}^n q(z_k)\ln q(z_k)$
\cite{jaynes,jaynes_2,skyrms,enk,cheeseman}. The intuitive rationale of
the method is that it provides the most unbiased choice of probability
compatible with constraints.

To find this relation, we
expand (\ref{6}) for a small $1-\beta$ (i.e.  $\beta\simeq 1$)
\BEA
\label{or}
L_\beta(\theta) &=& \sum_yp(y)\ln p_\theta(y)+\frac{1}{\beta}\sum_y p(y)\ln\left[ \sum_x p_\theta(x|y) e^{(\beta-1)\ln p_\theta(x|y)}  \right]\\
\label{ror}
&=& \sum_yp(y)\ln p_\theta(y)+(1-\beta)\sum_y p(y) S_y(\theta)+(1-\beta)^2\sum_y p(y) S_y(\theta) \\
\label{te}
&+&\frac{(1-\beta)^2}{2}\sum_{xy}p(y) p_\theta(x|y)\left(\, - S_y(\theta)+\ln p_\theta(x|y)  \, \right)^2
+{\cal O}\left((1-\beta)^3\right),\\
 S_y(\theta) &\equiv& -\sum_x  p_\theta(x|y)\ln  p_\theta(x|y),
\label{ga}
\EEA
where $S_y(\theta)$ is the entropy of $X$ for a fixed observation $Y=y$ and fixed parameters $\theta$.
When expanding $e^{(\beta-1)\ln p_\theta(x|y)}$ over ${(\beta-1)\ln
p_\theta(x|y)}$ we need to assume that $p_\theta(x|y)>0$, but eventually
a milder condition $p_\theta(x|y)\geq 0$ suffices because the terms in
(\ref{ror}--\ref{ga}) stay finite for $p_\theta(x|y)\to 0$. 

The zero-order term in $L_{\beta}(\theta)$ is naturally
$L(\theta)=\sum_yp(y)\ln p_\theta(y)$; see (\ref{ror}). But, as we
explained around (\ref{nono}), even when $N\gg 1$ in (\ref{glinka}), the
maximization of $L_1$ does not lead to a single result if the model is
not identifiable.  This degeneration will be (at least partially) lifted
if the next-order term $(1-\beta)\sum_y p(y) S_y(\theta)$ in $L_\beta$
is taken into account; cf.~(\ref{ror}). For $\beta<1$ this term will
tend to lift the degeneracy by selecting those maxima which achieve the
largest average entropy $\sum_y p(y) S_y(\theta)$. Hence for a small,
but positive $1-\beta$, the results of maximazing $\sum_yp(y)\ln
p_\theta(y)$ will (effectively) serve as constraints when maximizing
$\sum_y p(y) S_y(\theta)$.  
This is the relation between maximizing $L_\beta$
(for a small, positive $1-\beta$) and entropy maximization \footnote{Note
that the idea of lifting degeneracies of the maximum likelihood 
by maximizing the entropy over those degenerate solutions appeared
recently in the quantum maximum likelihood method \cite{hradil,singa}.
But there the degeneracies of the likelihood are due to incomplete
(noisy) data, i.e. they appear in a identifiable model.}. 

Note that when $p(y)$ converges to the true probabilities of $Y$, i.e.
when $N\gg 1$ in (\ref{glinka}), and when $\theta$ is fixed to its true
value, then $\sum_yp(y)S_y (\theta)$ is the conditional entropy of $X$
given $Y$ \cite{jaynes_2}.  The appearance of the conditional entropy is
reasonable given the fact that $Y$ is an observed variable. 

Within the second-order term ${\cal O}\left((1-\beta)^2\right)$ the
fluctuations of entropy enter into consideration: the degeneration will
be lifted by (simultaneously) maximizing the entropy variance and
maximizing the entropy; see (\ref{ror}, \ref{te}). 

Likewise, for $\beta>1$ (but $\beta\simeq 1$) the term $(1-\beta)\sum_y
p(y) S_y(\theta)$ in $L_\beta(\theta)$ predicts that among degenerate
maxima of $L_1(\theta)$, those of the minimal entropy will be selected.

\subsubsection{$Q$-function and generalized EM procedure}

$L_{\beta}(\theta)$ in (\ref{6}) admits a representation
via a suitably generalized $Q$-function, i.e.  its local maximum can be
calculated via the (generalized) expectation-maximization (EM) algorithm. Let
us define for two different values of $\theta$ and $\widetilde{\theta}$:
\BEA
\label{qq}
Q_\beta(\theta,\widetilde{\theta})=\sum_y p(y)\sum_x\zeta_\theta(x|y; \beta)\ln {p}_{\widetilde{\theta}}(x,y),
\EEA
where $\zeta_\theta(x|y; \beta)$ defined by (\ref{drud}) is formally a
conditional probability.  For $\beta=1$ we revert from (\ref{qq}) to the
average of the usual $Q$-function \cite{ephraim_review,wu}: $\sum_x p_\theta(x|y)\ln
{p}_{\widetilde{\theta}}(x,y)$, which is
the full log-likelihood $\ln {p}_{\widetilde{\theta}}(x,y)$ that
is averaged over the hidden variable $X$ given the observed $Y=y$ and calculated at
trial values $\theta$ and $\widetilde{\theta}$. 

Now non-negativity of the relative entropy:
\BEA
\label{qbit}
\frac{1}{\beta}\sum_y p(y)\sum_x \zeta_\theta(x|y; \beta)\ln\frac{\zeta_\theta(x|y;\beta)}{\zeta_{\widetilde{\theta}}(x|y;\beta)}\geq 0,
\EEA
implies after using (\ref{drud}, \ref{qq}) and re-arranging (\ref{qbit}):
\BEA
\label{boris}
L_\beta(\widetilde{\theta})-L_\beta({\theta})
\geq Q_\beta(\theta,\widetilde{\theta})-Q_\beta(\theta,{\theta}).
\EEA
Hence if for a fixed $\theta$ we choose $\widetilde{\theta}$ such that
$Q_\beta(\theta,\widetilde{\theta})>Q_\beta(\theta,{\theta})$, then this
will increase $L_\beta(\widetilde{\theta})$ over $L_\beta({\theta})$.
Eq.~(\ref{boris}) shows the main idea of EM: defining 
\BEA
\label{recu}
\theta_{k+1}={\rm argmax}_{\widetilde{\theta}} [\, Q_\beta(\theta_k,\widetilde{\theta})\,], 
\EEA
and starting from trial value $\theta_1$, we increase $L_\beta({\theta})$ sequentially, 
$L_\beta({\theta}_{k+1})\geq L_\beta({\theta_k})$, as (\ref{boris}) shows. 
Eq.~(\ref{qq}) implies:
\BEA
\label{godunov}
\left.\frac{\partial Q_\beta(\theta,\widetilde{\theta}) }{\partial \widetilde{\theta}}\right|_{\widetilde{\theta}=\theta}&=&
\frac{\partial L_\beta(\theta) }{\partial {\theta}}=
\sum_y \frac{p(y)}{\sum_{\bar{x}}p_\theta^\beta(\bar{x},y)  }\sum_x p_\theta^{\beta-1}(x,y)
\frac{\partial p_\theta(x,y) }{\partial {\theta}}.
\EEA
Eq.~(\ref{godunov}) shows that if we would find $\theta^*$ such that the maximum of
$Q_\beta(\theta^*,\widetilde{\theta})$ over $\widetilde{\theta}$ is
reached at $\widetilde{\theta}={\theta}^*$, i.e.
\BEA
\label{taras}
\left.\frac{\partial Q_\beta(\theta,\widetilde{\theta}) }{\partial \widetilde{\theta}}\right|_{\widetilde{\theta}=\theta=\theta^*}=
\left.\frac{\partial L_\beta(\theta) ) }{\partial {\theta}}\right|_{\theta=\theta^*}
=0,
\EEA
then $\theta^*$ can be a local maximum of $L_\beta(\theta)$, or an
inflection point of $L_\beta(\theta)$ (which has a direction along
which it maximizes), or|for a multidimensional $\theta$|a saddle point. Eq.~(\ref{taras})
holds if (\ref{recu}) converges. Thus similarly to the usual likelihood, $L_\beta(\theta)$
can be partially (i.e. generally not globally) maximized via (\ref{qq}).

\comment{
To employ (\ref{boris}, \ref{godunov}) in practice, one has to work with
a generalization of the Expectation-Maximization (EM) procedure
\cite{ephraim_review}, where starting from a trial value $\theta_0$, and
employing the above reasoning sequentially|i.e. using
$\theta_k=\widetilde{\theta}(\theta_{k-1})$ for $k\geq 1$, where
$\widetilde{\theta}(\theta_{k-1})$ maximizes
$Q_\beta(\theta_{k-1},\widetilde{\theta})$|one can really reach
$\theta^*$.  The convergence of this generalized EM is left as an open
issue; cf.~\cite{wu}. 
}

\subsection{First example (discrete random variables)} 
\label{simpo}

\subsubsection{Definition}

The following example is among simplest ones, but it does illustrate several 
general points of the approach based on maximizing $L_\beta(\theta)$. A binary random
variable $X$ ($x=\pm 1$) is hidden, while its noisy version $Y$ ($y=\pm 1$) 
is observed. The joint probability of $XY$ reads
\BEA
p_{gh}(x,y)=\frac{e^{gx+h xy}}{4\cosh h\cosh g},\quad
x=\pm 1,\quad y=\pm 1,
\label{wes}
\EEA
where $g>0$ and $h>0$ are unknown parameters: $g$ relates with the prior
probability of unobserved $X$, and $h$ relates to the noise. Since the
marginal probability of $Y$ holds:
\BEA
\label{pretoria}
p_{gh}(y)=p_z(y)=\frac{1}{2}(1+zy), \qquad z\equiv \tanh h\tanh g,
\EEA
even with infinite set of $Y$-observations one can determine only the
product $\tanh h\tanh g$, but not the separate factors $g$ and $h$. On
the other hand, the full model (\ref{wes}) is identifiable with respect
to $g$ and $h$, i.e. we have nonidentifiability in the sense of
(\ref{nono}, \ref{star}). Appendix \ref{bayo} discusses a Bayesian
approach to solving this nonidentifiability. As expected, if a good
(sharp) prior probabilities for $g$ or for $h$ are available, then the nonidentifiability
can be resolved.  However, when no prior information is available, one
is invited to employ noninformative priors \cite{jaynes}, which are improper for this
model, and which do not lead to any sensible outcomes; see Appendix \ref{bayo}. To the same end,
Appendix \ref{maxmayo} studies a decision-theoretic (maximin) approach
to this model, which also does not assume any prior information on $g$ and/or on $h$. This
approach also does not lead to sensible results. Thus, Appendices \ref{bayo}
and \ref{maxmayo} argue that the estimation of parameters in
(\ref{wes}) is a nontrivial problem.

We shall assume that a large ($N\gg 1$) set of observations is given in
(\ref{data}); hence $p(y)=p_z(y)$; see (\ref{6}, \ref{pretoria}). Omitting
irrelevant constants, we get from (\ref{6}, \ref{wes}, \ref{pretoria}):
\BEA
\label{fasol}
L_\beta(\hat g,\hat h)=-\ln\cosh \hat h-\ln\cosh \hat g
+\frac{1+z}{2\beta}\ln\cosh (\beta \hat g +\beta\hat h)+\frac{1-z}{2\beta}\ln\cosh (\beta \hat g -\beta\hat h),
\EEA
where $\hat g$ and $\hat h$ are estimates of (resp.) $g$ and $h$ to be
determined from maximizing (\ref{fasol}). Recall that we assumed $\hat
g>0$ and $\hat h>0$ as a prior information.  Eq.~(\ref{fasol}) is
invariant with respect to interchanging $\hat g$ and $\hat h$: $\hat
g\leftrightarrows \hat h$. 

\subsubsection{Solutions}

Now equations
$\partial L_\beta(\theta)/\partial \hat g=\partial L_\beta(\theta)/\partial \hat h=0$ reduce from (\ref{fasol}) to 
\BEA
\label{tutai}
\tanh (\beta\hat g+\beta \hat h)=\frac{\tanh \hat h+\tanh \hat g}{1+z}, \quad \tanh (\beta\hat g-\beta \hat h)=\frac{\tanh \hat h-\tanh \hat g}{1-z},
\EEA
where for $\beta=1$ we obtain from (\ref{tutai}) the expected $\tanh \hat h\tanh \hat g=z$. 
One can check that for $\beta<1$, the global maximum of (\ref{fasol}) is given by 
solutions of (\ref{tutai}), where 
\BEA
\label{gan}
&& \hat g=\hat h \quad {\rm hence} \\ 
&& \frac{1+z}{2}\tanh (2\beta\hat g)=\tanh \hat g,
\label{tugan}
\EEA
where (\ref{gan}) is a single maximum of the function (\ref{fasol}) that
has $\hat g\leftrightarrows \hat h$ symmetry.

For $\beta<1/2$ the only solution of (\ref{tugan}) is $\hat g=\hat h=0$,
which is far from holding $\tanh \hat h\tanh \hat g=z$; hence we
disregard the domain $\beta<1/2$.  For $\beta<1$, but $(1+z)\beta>1$,
there is a non-zero solution of (\ref{tugan}) that provides the global
maximum of $L_\beta$.  This solution is certainly better than the
previous $\hat g=\hat h=0$, but it also does not {\it exactly} hold the
constraint $\tanh \hat h\tanh \hat g=z$. This recovery|i.e. the
convergence $\hat g=\hat h\to {\rm arctanh} \sqrt{z}$|is achieved only
in the limit $\beta\to 1-$.  For any $\beta<1$ we thus have from
maximizing $L_\beta$: $\hat g=\hat h< {\rm arctanh} \sqrt{z}$.  Both
these facts are seen from (\ref{tugan}). 

The situation is different for $\beta>1$: under assumed $\hat g\geq 0$ and $\hat
h\geq 0$, we get two maxima of $L_\beta$ related by the transformation 
$\hat g\leftrightarrows \hat h$ to each other:
\BEA
\label{gle}
\hat g=\infty,~\hat h={\rm arctanh}z\qquad {\rm or}\qquad \hat h=\infty,~ \hat g={\rm arctanh}z.
\EEA
Both solutions hold $\tanh \hat h\tanh \hat g=z$; in a sense these
are the most extreme possibilities that hold this constraint
\footnote{Note that (\ref{gle}) can be obtained in a more artificial
way, by replacing $x\to x^\circ(y;g,h)\equiv\sum_x
xp_{gh}(x|y)=\tanh(g+hy) $ in $p_{gh}(x,y)$, and then maximizing
$\sum_{y} p_{gh}(y) \ln p_{gh}(x^\circ(y;g,h),y)$ over $g$ and
$h$; cf. this procedure with (\ref{udodo}).  Replacing $x\to
x^\circ(y;g,h)$ is formal, since $p_{gh}(x,y)$ is (strictly speaking)
not defined for a real $x$. Still for this model this formal procedure
leads to (\ref{gle}).}. 

We emphasize that one does not need to focus exclusively on maximizing
$L_\beta(\hat g,\hat h)$ over $\hat g$ and $\hat h$. 
We note that $\int \d g\,\d h \,e^{L_\beta( g, h)}$ is finite and hence we
can consider $e^{L_\beta(\hat g,\hat h)}/\int \d g\,\d h\, e^{L_\beta( g,
h)}$ as a joint density of $\hat g$ and $\hat h$, which is still symmetric
with respect to $\hat g\leftrightarrows \hat h$. 

\subsubsection{Overconfidence}

Returning to solutions (\ref{tugan}) and (\ref{gle}), let us argue that
there is a sense in which (\ref{tugan}) is better than (\ref{gle}). To
this end, we should enlarge our consideration and ask which solution is
more suitable from the viewpoint of finding an estimate $\hat x(y)$ of
the hidden variable $X$ given the observed value of $Y=y$. This
estimation can be done via maximizing the overlap (or the risk
function): $O(y; \hat g, \hat h)=\sum_{x=\pm 1} \hat x (y)x p_{\hat
g\hat h}(x|y)$ over $\hat x(y)$; see (\ref{wes}). The maximization
produces $\hat x(y)={\rm sign}[\hat g+\hat hy]$, and the quality of the
estimation can be judged via the average overlap [cf.~(\ref{pretoria})]:
\BEA
\label{tartar}
\bar{O}(z; \hat g, \hat h)=\sum_{y=\pm 1}p_{gh}(y)O(y; \hat g, \hat h)
=\frac{1+z}{2}\tanh\left|\hat g+\hat h\right|+\frac{1-z}{2}\tanh\left|\hat g-\hat h\right|.
\EEA
\comment{The value of (\ref{gle}) reads for $\hat g>\hat h$ 
(if $\hat g<\hat h$, then the vaues of $\hat g$ and $\hat h$ 
are interchanged): $\frac{\tanh \hat g\,(1-\tanh^2\hat h)+
z \tanh \hat h\,(1-\tanh^2\hat g)}{1-\tanh^2 \hat g\,\tanh^2\hat h }$. }
If the values of $g$ and $h$ are known precisely, $g=\hat g$ and
$h=\hat h$, then together with $z= \tanh h\tanh g$ we get from (\ref{tartar}):
$\bar{O}(z; g, h)={\rm max}[\tanh g, \tanh h]$.  Now employing in
(\ref{tartar}) solution (\ref{gle}), we get $\bar{O}(z; \hat g, \hat h)
=1 >\bar{O}(z; g, h)$.  This {\it overconfidence} is not desirable,
because with approximate values of parameters we do not expect to have a
better estimation quality than with the true values. In contrast, using
(\ref{tugan}) in (\ref{tartar}) we get a reasonable conclusion:
\BEA
\bar{O}(z; \hat g, \hat h)=\frac{1+z}{2}\tanh 2\hat g< \sqrt{z} <\bar{O}(z; g, h). 
\EEA
Hence, from this viewpoint, 
the best regime is $\beta\lesssim 1$, since we approximately hold the contraint
$\tanh\hat h \tanh\hat g=\tanh h \tanh g$, and also $\bar{O}(z; \hat g, \hat h)
<\bar{O}(z; g, h)$. Moreover, the $\beta\lesssim 1$-solution is unique in contrast
to (\ref{gle}).

\subsection{Second example (continuous random variables)}
\label{counter}

While the previous example showed that the maximization
of $L_{\beta<1}$ can produce reasonable results, here
we discuss a continuous-variable example, where the similar
maximization leads nowhere without additional assumptions 
on the model. Consider an analogue of (\ref{wes}):
\BEA
p_{gh}(x,y)=g\, e^{-gx}\, h\,x\,e^{-h xy},\quad
x\geq 0,\quad y\geq 0, \quad g>0, \quad h>0,
\label{weso}
\EEA
where $X$ (hidden) and $Y$ (observed) are nonnegative, continuous 
random variables, while $g$ and $h$ are positive unknown parameters. 
The full model is identifiable; e.g. the maximum-likelihood estimates
of $g$ and $h$ read (resp.): $1/x $ and $1/(xy)$, where $x$ and $y$
are observed values of $X$ and $Y$. But the marginal model is not
identifiable, since
\BEA
p_{gh}(y)=p_{\chi}(y)=\frac{\chi}{[y+\chi]^2},\qquad \chi\equiv g/h,
\EEA
depends on the ratio $\chi$ of two unknown parameters; cf.
(\ref{pretoria}). Maximizing over $\hat\chi$ the marginal likelihood
$\int_0^\infty \d y\, p_{\chi}(y)\ln p_{\hat\chi}(y)$|for a large $N\gg
1$ number of observations in (\ref{data})|leads to the correct outcome
$\hat\chi=\chi$.  But the individual values of unknown parameters $\hat g$
and $\hat h$ are not determined in this way. 

We now employ (\ref{6}, \ref{weso}) with an obvious generalization of
(\ref{6}) to continuous random variables, and write for $L_\beta(\hat
g,\hat h)$ (again assuming $N\gg 1$):
\BEA
\label{nora}
L_\beta(\hat g,\hat h)=
\frac{1}{\beta}\int_0^\infty \d y\, p_{\chi}(y)\ln\int_0^\infty \d x\,p_{\hat g \hat h}^\beta(x,y)
&=&\frac{1}{\beta}\ln [\,\Gamma(\beta)/\beta\,]+(1-\frac{1}{\beta})\ln\hat h \\
&+&\ln (\hat \chi)
-\frac{\beta+1}{\beta}\,\,\frac{\chi\ln [\chi] - \hat \chi\ln [\hat \chi]}{\chi-\hat\chi},
\label{shen}
\EEA
where $\Gamma(\beta)$ is the Euler's Gamma-function, and where $\hat\chi\equiv \hat g/\hat h$.
It is seen that $L_\beta$ expresses in terms of two unknown parameters: $\hat h$ and $\hat\chi$.
Hence the maximization of $L_\beta$ can be carried out independently over $\hat h$ and $\hat\chi$.
Now the maximization of (\ref{shen}) over $\hat\chi$ produces for a fixed $\hat h$ a finite outcome
for $\hat\chi$ (see below), while the maximization of (\ref{nora}) over $\hat h$ leads to $\hat h\to 0$
for $\beta>1$ and to $\hat h\to\infty$ for $\beta<1$. Hence $L_{\beta\not=1}(\hat g,\hat h)=
L_{\beta\not=1}(\hat \chi,\hat h)$ does not have
maxima for positive and finite $\hat g$ and $\hat h$, as required for having a reasonable model in 
(\ref{weso}). Note that this situation is worse than the maximization
of the marginal likelihood $L_1$, because there at least the value of the ratio $\hat\chi=\chi$ was recovered
correctly (in the limit of infinite number of observations).

The situation with maximizing $L_{\beta<1}(\hat \chi,\hat h)$ in (\ref{nora}, \ref{shen}) improves,
if we assume an additional prior information on $h$:
\BEA
\label{ko}
h\leq H,
\EEA
where $H>0$ is a new {\it and known} parameter. Now (\ref{nora}, \ref{shen}) is to be maximized over
$\hat \chi$ and over $\hat h$ under constraint $\hat h\leq H$. For $\beta<1$ this maximization produces 
reasonable results:
\BEA
&&{\rm argmax}_{\hat h,\hat \chi} [\, L_{\beta<1}(\hat \chi,\hat h)\,]=\left(\, {\hat h=H, \hat \chi=f_\beta(\chi)}\,\right),\\
&& f_\beta(\chi)<\chi\quad {\rm for} \quad {\beta<1},\quad {\rm and}\quad f_\beta(\chi)\to \chi\quad {\rm for} \quad {\beta\to 1}.
\EEA
I.e. for $\beta\to 1$, but $\beta<1$ we a unique maximization outcome: 
$\hat h=H$ and $\hat g=H \chi$. Note that the maximization
of $L_{\beta>1}$ is still not sensible, since it leads to $\hat h\to 0$. 

To conclude this continuous-variable example, here the maximization of
$L_{\beta<1}$ produces unique and correct results for unknown parameters
$\hat g$ and $\hat h$ (correct in the sense of reproducing the ratio
$g/h$), at the cost of additional assumption (\ref{ko}). If this 
assumption is not made, then only the maximization of $L_{\beta=1}$,
i.e. of the usual marginal likelihood, is sensible for this model.  The 
maximization of $L_{\beta>1}$ is never sensible here. 

\comment{
Now $\partial_{\hat g} L_\beta(\hat g,\hat h)=0$ and $\partial_{\hat h}
L_\beta(\hat g,\hat h)=0$ reduce to two different equations for a single
quantity $\hat g/\hat h$, which are not satisfied together for
$\beta\not=1$. 
}

\section{Mixture model with unknown probabilities }
\label{mixo}


Now we focus on a sufficiently general mixture model, which will
allow us to study in detail the structure of $L_\beta$ and its
dependence on $\beta$. In mixture model (\ref{1}) probabilities 
$p(x)$ and $p(y|x)$ are unknown. The prior information on them is 
introduced below. We shall skip $\theta$ and denote unknown probabilities by hats:
\BEA
\label{bilibin}
\hat p(x,y)=\hat p(x) \hat p(y|x).
\EEA
Then $L_{\beta}(\theta)$ reads from (\ref{6})
\BEA
\label{gosh}
L_{\beta}=
\frac{1}{\beta}\sum_yp(y)\ln\left[\sum_x \hat p^\beta(x,y) \right].
\EEA
If $N\gg 1$ in (\ref{data}), and hence frequencies $p(y)$ converged to probabilities of $Y$,
quantities in (\ref{bilibin}) have to hold:
\BEA
\label{co}
\label{5}
\sum_x\hat p(x,y)=p(y),
\EEA
which is also produced by the maximization of $L_1$ from (\ref{gosh}). 
Eq.~(\ref{5}) has $m-1$ known quantities $p(y_1),...,p(y_m)$ (note the
constraint $\sum_{i=1}^m p(y_i)=1$). If all $\hat p(x)$ and $\hat p(y|x)$
are unknown (apart of holding (\ref{5})), then we have
$nm-m$ unknown variables: $nm-1$ parameters $\hat p(x,y)$ minus $m-1$ known
parameters $p(y)$. Already for $n=2$, $nm-m$ is larger than the number
$m-1$ of known variables. As expected, (\ref{5}) will not give a unique
solution, and the model is nonidentifiable; cf.~(\ref{nono}).

Apart of (\ref{co}), further constraints are also possible. Such constraints amount
to various forms of prior information; e.g. $\hat p(x)$ and $\hat p(y|x)$ hold
a linear constraint:
\BEA
\label{beri2}
\sum_{xy}E(x,y) \hat p(x,y)=E,
\label{tari2}
\EEA
where $E(x,y)$ is some function of $x$ and $y$ with a known average $E$.
For instance, $E(x,y)=xy$ refers to the correlation between $X$ and $Y$. 
Another example of (\ref{beri2}) is when one of probabilities $\hat p(x,y)$
is known precisely. Note that several linear constraints can be implemented
simultaneously, this does not increase the analytical difficulty of treating
the model. Constraints similar to (\ref{beri2}) decrease the
number of (effectively) unknown variables, but we shall focus on the
situation, where they cannot select a single solution of (\ref{co}), i.e. the
nonidentifiability is kept.

Once the maximization of $L_1$ does not lead to any definite outcome, we
look at maximizing $L_\beta$. To this end, it will be useful to recall
the concavity of $L_{\beta\leq 1}$; cf.~(\ref{glinka}). The advantage of linear
constraints [cf.~(\ref{co}, \ref{beri2})], is that unknown $\hat p(x,y)$ are
defined over a convex set. Eq.~(\ref{glinka}) means that for $\beta<1$
there can be only a single internal (with respect to the convex set)
point $p_0$, where the gradient of $L_{\beta\leq 1}(p)$ vanishes, $\nabla
L_{\beta\geq 1}|_{\hat p=\hat p_0}=0$, and $\hat p_0$ is the global maximum
of $L_{\beta<1}(\hat p)$. 

\comment{
The concavity of $L_{\beta\leq 1}$ can be confirmed its Hessian 
(calculated without imposing restrictions on $\hat p(x,y)$): 
\BEA
\label{taron}
&&\frac{\partial^2 L_\beta}{\partial \hat p(x,y) \partial \hat p(x',y')}=A_{xy,\,x'y'} - B_{xy,\,x'y'}, \\
&&A_{xy,\,x'y'}\equiv p(y)
\frac{(\beta-1)\hat p^{\beta-2} (x,y)\,\delta_{xx'}\delta_{yy'}}{\sum_{\bar{x}} \hat p^\beta(\bar{x},y)}, \qquad
B_{xy,\,x'y'}\equiv p(y)
\frac{\beta\hat p^{\beta-1} (x,y)\, \hat p^{\beta-1} (x',y)\,\delta_{yy'}}{\left[\sum_{\bar{x}} \hat p^\beta(\bar{x},y)\right]^2}.
\label{tarson}
\EEA
Now $B_{xy,\,x'y'}$ is nonnegative definite (i.e. its eigenvalues are
not negative). It can have zero eigenvalues for eigenvectors $u(x,y)$
that hold $\sum_x \hat p^{\beta-1}(x,y) u(x,y)=0$ for all values of $y$.
For $\beta<1$, $A_{xy,\,x'y'}$ is negative definite. Hence we cofirm
that $L_{\beta\leq 1}$ is concave, since the eigenvalues of
(\ref{taron}) are not positive for $\beta<1$.  And it will stay concave
after restricting its domain to a narrower convex set defined by
(\ref{co}, \ref{beri2}). 

The situation is different for $\beta>1$, since $A_{xy,\,x'y'}$ and
$B_{xy,\,x'y'}$ do not (generally) commute, and $A_{xy,\,x'y'}$ is
nonnegative definite. Hence (\ref{taron}) is generally nonnegative
definite in the zero-eigenvalue subspace of $B_{xy,\,x'y'}$, i.e.
$L_{\beta>1}$ cannot have a local maxima in that subspace. 
}

\section{Maximizing the generalized likelihood for $\beta\leq 1$}
\label{beta<1}

\comment{

\subsection{Complete ignorance}
\label{igno}

Let us now assume that in (\ref{6}) all $\hat p(y|x)$ and $\hat p(x)$ are unknown. 
We introduce constraint (\ref{co}) into (\ref{6}) treating the Lagrange function
\BEA
\label{7}
{\cal L}_\beta=\frac{1}{\beta}\sum_yp(y)\ln\left[\sum_x \hat p^\beta(x,y)\right]-\sum_{xy}\Gamma(y)\hat p(x,y),
\EEA
where $\Gamma(y)$ are Lagrange multipliers. We get
\BEA
\label{7.77}
\frac{\partial {\cal L}_\beta}{\partial \hat p(x,y)}=p(y)\,\frac{\hat p^{\beta-1}(x,y)}{\sum_{x'} \hat p^{\beta}(x',y)}-\Gamma(y)=0,
\EEA
which leads to 
\BEA
\label{8}
p(y)\hat p^\beta(x,y)=\Gamma(y)\hat p(x,y)\sum_{x'}\hat p^\beta(x',y).
\EEA
Summing (\ref{8}) over $x$ we get $\Gamma(y)=1$. Then dividing (\ref{8}) over $\hat p(x,y)$, and 
noting that the resulting right-hand-side does not depend on $x$ we get
\BEA
\label{9}
\hat p(x,y)=p(y)\frac{1}{n}.
\EEA
This is a reasonable solution of the problem, given the fact
that we know nothing about $\hat p(x)$ and $\hat p(y|x)$ besides 
constraints (\ref{co}); cf.~(\ref{ushi}). Note that we will get 
the same result (\ref{9}), even if we do not impose the contraint (\ref{co}) (but still impose $\sum_{xy}\hat p(x,y)$, of course).

We get from (\ref{9}, \ref{taron}) for the Hessian matrix that determines the stablity of (\ref{9}):
\BEA
\label{brab}
\frac{\partial^2 {\cal L}_\beta}{\partial \hat p(x,y)\,\partial \hat p(x',y)}
=\frac{n}{p(y)}\left( (\beta-1)\,\delta_{xx'}-\frac{\beta}{n} \right).
\EEA
To find eigenvalues of (\ref{brab}), recall that a square matrix $\{A_{ij}=\alpha\delta_{ij}+\beta\}_{i,j=1}^n$ has two eigenvalues: $\alpha$ is $n-1$
times degenerate and refers to eigenvectors that hold $\sum_i a^{[1]}_i=0$. The other eigenvalue $a+bn$ is nondegenerate and has the eigenvector with
$a^{[2]}_i=a^{[2]}$. We conclude that (\ref{brab}) has two eigenvalues $-1$ (nondegenerate) and $\frac{n(\beta-1)}{p(y)}$ ($n-1$-times degenerate). 
Hence for $\beta<1$ the solution (\ref{9}) is stable, i.e. it refers to the maximum of $L_\beta$. 
}

\subsection{Known probability of $X$ }
\label{margo}

As the first exercise in maximizing $L_{\beta<1}$ for the present model,
let us assume that (prior) probabilities $p(x)$ are known. Hence
\BEA
\label{op}
p(x)=\sum_y\hat p(x,y).
\EEA 
The Lagrange function reads:
\BEA
{\cal L}_\beta=
\frac{1}{\beta}\sum_y p(y)\ln\left[\sum_x \hat p^\beta(x,y)  \right]
-\sum_{xy}\gamma(x)\hat p(x,y),
\EEA
where $\gamma(x)$ are Lagrange multipliers of
(\ref{op}). Now $\frac{\partial {\cal L}_\beta }{\partial\hat p(x,y) }=0$
amounts to 
\BEA
\label{ev}
p(y)\frac{\hat p^{\beta-1}(x,y)   }{\sum_{\bar{x}} \hat p^\beta(\bar x,y) }=\gamma(x).
\EEA
Since the right-hand-side of (\ref{ev}) does not depend on $y$ so should its left-hand-side, which is only possible under
\BEA
\label{unco}
\hat p(x,y)=p(y)p(x).
\EEA
Once (\ref{unco}) solves (\ref{ev}), it is the global maximum of
$L_{\beta<1}$, since the latter is concave.  Recall that $p(y)$ are
generally the observed frequencies of (\ref{data}). Though (\ref{unco})
may not very useful by itself, it still shows that maximizing
$L_{\beta<1}$ under (\ref{op}) leads to a reasonable null model in a
nonidentifiable situation. Imposing other constraints on $\hat p(x,y)$
does lead to nontrivial predictions, as we now proceed to show.

\comment{
Note that (\ref{unco}) can be also 
obtained if we under constraints (\ref{op}) we maximize the entropy
$-\sum_{xy}\hat p(x,y)\ln \hat p(x,y)$; cf. section \ref{uraa}.}

\subsection{Known average}
\label{oslo}

\subsubsection{Derivation}

Let us turn to maximizing $L_\beta$ under constraint (\ref{beri2}).
The Lagrange function reads:
\BEA
{\cal L}_\beta=
\frac{1}{\beta}\sum_yp(y)\ln\left[\sum_x \hat p^\beta(x,y) \right]
- \delta\sum_{xy}\hat p(x,y)
-\gamma \sum_{xy}E(x,y)\hat p(x,y),
\EEA
where $\delta$ refers to the normalization $\sum_{xy} \hat
p(x,y)=1$ and $\gamma$ enforces (\ref{beri2}).
Now $\frac{\partial {\cal L}_\beta }{\partial\hat p(x,y) }=0$ leads to 
\BEA
\label{mur2}
\frac{p(y) \hat p^{\beta-1}(x,y)   }{\sum_{\bar{x}} \hat p^\beta(\bar x,y) }=\delta+\gamma E(x,y),
\EEA
which is solved as 
\BEA \label{mur1}  
\hat p(x,y)=p(y)\,\frac{\left[\delta+\gamma E(x,y)
\right]^{\frac{1}{\beta-1}}} {\sum_{\bar{x}} \left[\delta+\gamma E(\bar
x,y) \right]^{\frac{\beta}{\beta-1}} }, 
\EEA
where $\gamma$ and $\delta$ are found from the normalization and from (\ref{beri2}):
\BEA
\label{khr0} 
&& \gamma =\frac{1}{E}\left(1-\delta \right),\\ 
&& \sum_{y}\hat p(y)=1, \quad \hat p(y)\equiv
p(y) \, \frac{\sum_x\left[\delta+\gamma E(x,y)
\right]^{\frac{1}{\beta-1}}} {\sum_{\bar{x}} \left[\delta+\gamma
E(\bar x,y) \right]^{\frac{\beta}{\beta-1}} }.
\label{khr1} 
\EEA 
Note that (\ref{khr0}, \ref{khr1}) have a spurious solution $\delta=1$, 
which is to be avoided in numerical determination of $\delta$.

\subsubsection{Features of (\ref{mur1}--\ref{khr1})}

{\bf 1.} Constraint (\ref{beri2}) is invariant with respect to
multiplying $E(x,y)$ and $E$ by a number. Hence $\hat p(x,y)$ in
(\ref{mur1}) is also invariant to this transformation, as seen from
(\ref{mur1}, \ref{khr0}), where $\delta$ and $\gamma E$ do not change
after multiplication. 

Constraint (\ref{beri2}) is also invariant with respect to 
shifting $E(x,y)$ and $E$ by a constant factor $a$: $E'(x,y)=E(x,y)+a$
and $E'=E+a$. Hence we can always choose $E(x,y)>0$ and $E>0$.
Now $\hat p(x,y)$ in (\ref{mur1}) is also invariant under this transformation, because
\BEA
\label{nush}
\delta+(1-\delta)\frac{E(x,y)}{E}=\delta'+(1-\delta') \frac{E'(x,y)}{E'},
\EEA
due to
\BEA
\gamma' =\frac{1}{E'}\left(1-\delta' \right), \qquad \delta'=\delta(1+\frac{a}{E})-\frac{a}{E}.
\EEA

{\bf 2.} Eq.~(\ref{mur1}) predicts independent variables $X$ and
$Y$, if $E(x,y)$ does not depend on $y$; i.e. having no prior
information on the dependency between $X$ and $Y$ leads to predicting them
to be independent \cite{jaynes}. This feature can be generalized 
showing that $\hat p(x,y)$ predicted by
(\ref{mur1}) is not more precise than $E(x,y)$:
assume that the range of $y$ is divided into mutually exclusive domains
${\cal S}_1,..., {\cal S}_M$, so that $E(x,y)=E_m(x)$ whenever $y\in
{\cal S}_m$.  Now denoting $p_m={\rm Pr}(y\in {\cal S}_m)=\sum_{y\in
{\cal S}_m}p(y)$ and $\hat p_m(x)= {\rm Pr}(x,\, y\in {\cal
S}_m)=\sum_{y\in {\cal S}_m}\hat p(x,y)$, we get that the shape of
(\ref{mur1}) coarse-grains and stays invariant: \BEA \label{surok2} \hat
p_m(x)=p_m\,\frac{\left[\delta+\gamma E_m(x)
\right]^{\frac{1}{\beta-1}}} {\sum_{\bar{m}=1}^M \left[\delta+\gamma
E_{\bar m}(x) \right]^{\frac{\beta}{\beta-1}} }, \qquad m=1,...,M.  \EEA

\begin{figure*}[t] 
\includegraphics[width=7cm]{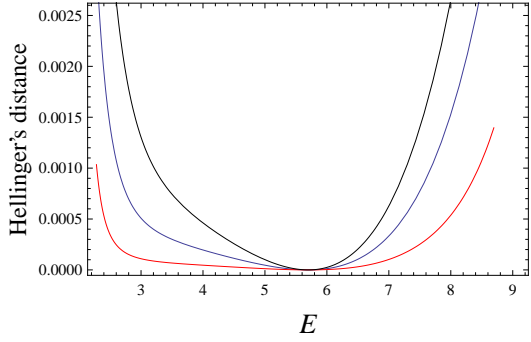} 
\caption{
Hellinger's distance $1-\sum_{k=1}^4\sqrt{\hat p(y_k) p(y_k)}$ between
$\hat p(y)=\sum_x\hat p(x,y)$ from (\ref{mur1}) and $p(y)$
for $n=m=4$, $p(y)=(0.4, 0.01, 0.5, 0.09)$, $E(x_k,y_l)=kl$ ($k,l=1,..,4$)
and various values of $E$ that hold (\ref{turku}). From 
top to bottom: $\beta=0.85$ (black curve), $\beta=0.9$ 
(blue curve) and $\beta=0.95$ (red curve). } 
\label{fig0}
\end{figure*}

{\bf 3.} We emphasize that the marginal probability $\hat
p(y)=\sum_x\hat p(x,y)$ from (\ref{mur1}) is generally not equal to
$p(y)$, i.e. (\ref{5}) does not follow from (\ref{mur1}). Now $\hat
p(y)\not=p(y)$ is not prohibited, if $p(y)$ are finite-sample
frequencies. But when $N\gg 1$ in (\ref{data}), then $\hat p(y)=p(y)$ is
demanded. This equality can be imposed via constraints |additional to
(\ref{beri2})|and this will lead to a joint probability different from
(\ref{mur1}); see Appendix \ref{two} for details.  Instead of imposing
additional constraints, we note from (\ref{mur1}, \ref{khr0}) that for
$\beta<1$ and $\beta\simeq 1$ (written together as $\beta\lesssim 1$),
we get $\delta\to 1$, and $\hat p(x,y)$ simplifies as
\BEA 
\hat p(x,y)&=& p(y)\,\frac{\left[1+\delta-1+\frac{1-\delta}{E} \,E(x,y)
\right]^{\frac{1}{\beta-1}}} {\sum_{\bar{x}} \left[1+\delta-1+\frac{1-\delta}{E} \,E(\bar
x,y) \right]^{\frac{\beta}{\beta-1}} }\nonumber\\
\label{mur11}  
&\simeq& p(y)\,\frac{  e^{-\Gamma E(x,y)}   }{ \sum_{\bar{x}} e^{-\Gamma E(\bar{x},y)}   }, \\
\Gamma&\equiv&\frac{1}{E}\,\frac{1-\delta}{1-\beta},
\label{mur12}  
\EEA
where $\Gamma$ stays finite in the limit $\beta\to 1-0$. It is clear from (\ref{mur11}, \ref{mur12}) that in this limit 
(\ref{5}) does follow from (\ref{mur1}): $\hat p(y)= p(y)$; cf.~section \ref{simpo}.

For the present analytically solvable situation, we able to take the the
limit $\beta\to 1-0$ and deduce (\ref{mur11}, \ref{mur12}). However,
upon more general usage of $L_\beta$ (and its maximization) this will
not be possible, since taking $\beta\approx 1$ in $L_\beta$ will run
into problems inherited from $L_1$ (quasi-degeneracy of maxima {\it
etc}). Hence it is important to know how close $\beta$ should be to $1$
for recovering $\hat p(y)\simeq p(y)$.  Fig.~(\ref{fig0}) illustrates 
this question by looking at Hellinger's distance between $\hat
p(y)$ and $p(y)$. It is seen that $0.9\leq \beta<0.95$ is already
sufficient for getting $\hat p(y)\simeq p(y)$ sufficiently precisely for
almost all values of $E$. 

{\bf 4.} Here are finally certain subsidiary, but useful features. 
When $p(y)$ are the true probabilities of $Y$, then $E$ is supposed to hold 
the following constraints:
\BEA
\label{turku}
&& \sum_y p(y) E(\widetilde{x}(y),y)\leq E\leq \sum_y p(y) E(\hat{x}(y),y), \\
&& \widetilde{x}(y)\equiv {\rm argmin}_x[\,E(x,y)\,], \qquad
\hat{x}(y)\equiv{\rm argmax}_x[\,E(x,y)\,].
\label{turku2}
\EEA
In addition, there is a relation that can be deduced directly from (\ref{mur11}, \ref{mur12}),
but appears to hold more generally, i.e. also for $\beta<1$:
\BEA
{\rm sign}\left[\gamma\right]={\rm sign}\left[ \frac{1}{n}\sum_{xy}p(y)E(x,y)  -E\right]. 
\label{grosh}
\EEA

\comment{
{\bf To be reconsidered:}{\it 
It should be clear that the maximizing (\ref{6}) is a more general
procedure, because it can be applied to those cases when the data
(\ref{data}) is not large, i.e. $N\gg 1$ does not hold. Hence the exact
probabilities $\hat p(y)$ are not known, and one should work with
(\ref{44}) instead of (\ref{4}). In other words, for $N\not\gg 1$
parameters are not probabilities, i.e. the entropy for them is not
defined \footnote{I am not sure that this argument is really conclusive.
One should keep in memory that there can be generalized maximum entropy
methods that we compare with the maximizing of (\ref{6}); in this
context see \cite{singa}.}. } }

\section{Numerical comparison with random choices of nonidentifiable parameters}
\label{compo}

\begin{table}
\begin{center}
\caption{The values of $D_1$ and $D_2$ given by (resp.) (\ref{d1}) and
(\ref{d2}) for $x=1,..,n$ and $y=1,...m=n$ and $E_1(x,y)=|x-y|$, $E_2(x,y)=xy$.
The averaging in (\ref{d1}, \ref{d2}) was taken over $S=10^3$ samples. We took $\beta=0.95$.
For completeness, we also presented the analogues of $D_1$ and $D_2$ (denoted by $K_1$ and $K_2$, respectively), where
Hellinger's distance in (\ref{hellinger}) is replaced by the relative entropy:
${\rm dist}[\pi_k,\hat p_k]\to \sum_{xy} \pi_k(x,y)\ln\frac{\pi_k(x,y)}{\hat p_k(x,y)} $.
Both choices support the same conclusion: $D_1<D_2$, $K_1<K_2$.}
\begin{tabular}{||c||c|c||}
\hline\hline
             & $n=m=4$  & $n=m=5$  \\
\hline\hline
  $E_1$   & $D_1=0.041$, $D_2=0.092$~  & ~$D_1=0.046$, $D_2=0.098$ \\
          & $K_1=0.145$, $K_2=0.410$~  & ~$K_1=0.153$, $K_2=0.442$                     \\
\hline
  $E_2$   & $D_1=0.041$, $D_2=0.089$~  & ~$D_1=0.045$, $D_2=0.096$ \\
          & $K_1=0.143$, $K_2=0.434$~  & ~$K_1=0.155$, $K_2=0.430$                     \\
\hline\hline
\end{tabular}
\end{center}
\label{tab1}
\end{table}

In this section we compare predictions obtained from maximizing
$L_{\beta<1}$ with the standard attitude of practitioners towards
nonidentifiability: people either take a maximum of the (marginal)
likelihood $L_1$, postulating that if there are many maxima, they are
eventually equivalent. Or, within a more careful, but also more
laborious approach, they average over sufficiently many such maxima. For
the studied model these maxima are given by (\ref{5}), and the
comparison will show that maximizing $L_{\beta\lesssim 1}$ is superior
with respect to such random selection methods. 

Let us assume that we know the true joint probability $\pi_k(x,y)$ of
$X$ and $Y$ (the meaning of an integer $k$ is specified below). Given
$\pi_k(x,y)$ and $E(x,y)$ we calculate the marginal probability of $Y$ and
the constraint
\BEA
\label{boom}
p_k(y)=\sum_x \pi_k(x,y), \qquad E_k=\sum_{xy}E(x,y)\pi_k(x,y).
\EEA
Using (\ref{mur1}--\ref{khr1}), and $p_k(y)$ and $E_k$ from (\ref{boom}) 
we recover $\hat p_k(x,y)$ that depends on $\beta<1$.
Recalling the discussion around (\ref{mur12}), we shall work with $\beta=0.95$.

The quality of $\hat p_k(x,y)$|given by (\ref{mur1}--\ref{khr1}, 
\ref{boom}) as a solution to the problem of estimating $\pi_k(x,y)$|can 
be judged from the distance ${\rm dist}[\pi_k,\hat p_k]$, which (for clarity)
is chosen to be Hellinger's distance between two probabilities:
\BEA
{\rm dist}[\pi_k,\hat p_k]\equiv 1-\sum_{xy}\sqrt{\hat p_k(x,y)\, \pi_k(x,y)}.
\label{hellinger}
\EEA
Now (\ref{hellinger}) depends on the choice of $\pi_k(x,y)$.
To make this dependence weaker, i.e. to make the situation less subjective, we
assume that $\pi_k(x,y)$ for $k=1,...S\gg 1$ are generated randomly and
independently from each other. The simplest possible mechanism suits
our purposes: we choose $\Pi_k(x,y)$ as $n\times
m\times S$ independent random variables homogeneously distributed
in $[0,A]$ (the choice of $A$ does not seriously influence on the
situation provided that $A\geq 1$), and then calculate:
\BEA
\label{ortiz}
\pi_k(x,y)=\Pi_k(x,y)\left/{\sum}_{\bar x \bar y}\right. \Pi_k(\bar x,\bar y). 
\EEA
Thus for $S\gg 1$ we define from (\ref{hellinger}) the averaged distance:
\BEA
\label{d1}
D_1=\frac{1}{S}\sum_{k=1}^S {\rm dist}[\pi_k,\hat p_k],
\EEA
which estimates the quality of $\hat p(x,y)$ in predicting the
(known) joint probability. To comment on the above choice $\beta=0.95$,
we note that from our numeric results that the dependence of $D_1$ on
$\beta$ is anyhow weak, e.g. it typically changes by 1 \% when changing $\beta$
from $0.7$ to $1$. 

Now $D_1$ will be compared with the situation, where|given $p_k(y)$ and
$E_k$ from (\ref{boom})| we do not employ (\ref{mur1}--\ref{khr1}), but
instead guess the joint probability of $X$ and $Y$. This will be done by
picking up randomly|via the same mechanism, as in (\ref{ortiz})|a
conditional probability $\widetilde{p}_k(x|y)$, with an additional
condition that it holds $\sum_{xy}p_k(y)\widetilde{p}_k(x|y)E(x,y)=E_k$
\footnote{In more detail, this goes as follows: given $\pi_k(x,y)$ we
find $E_k$ and $p_k(y)$ via (\ref{boom}). Next for a fixed $k$ we
randomly generate $nm-1$ positive variables
$\{\widetilde{\Pi}_k(x,y)\}$; their number is $nm-1$, since
$\widetilde{\Pi}_k(n,m)$ is absent.  Then we look at equation
(\ref{boom}): $\sum_yp(y)\frac{\sum_{x}
\widetilde{\Pi}_k(x,y)[E(x,y)-E_k]}{\sum_{\bar{x}}
\widetilde{\Pi}_k(\bar{x},y)}=0$, with unknown $\widetilde{\Pi}_k(n,m)$.
If this equation is solved with a nonnegative solution
$\widetilde{\Pi}_k(n,m)$, the latter is joined to
$\{\widetilde{\Pi}_k(x,y)\}$, and we take
$\widetilde{p}_k(x|y)=\widetilde{\Pi}_k(x,y)\left/{\sum}_{\bar x}\right.
\widetilde{\Pi}_k(\bar x, y)$ as the sought random conditional
probability. Otherwise, if the equation is not solved with a positive
$\widetilde{\Pi}_k(n,m)$, we generate $\{\widetilde{\Pi}_k(x,y)\}$ anew,
till $\widetilde{\Pi}_k(n,m)>0$.}; see (\ref{boom}). Thereby we construct 
\BEA 
\label{d2}
D_2=\frac{1}{S}\sum_{k=1}^S {\rm dist}[\pi_k, p_k(y) \widetilde{p}_k(x|y)].  
\EEA
Due to $S\gg 1$ in (\ref{d2}), $D_2$ is (almost) a sure quantity. 
Table~I compares $D_2$ with $D_1$ for a representative set of
parameters. It is seen that $D_2$ is some two times larger than $D_1$,
i.e. a random solution is worse than (\ref{mur1}--\ref{khr1}). Table~I
also shows that $D_2>D_1$ holds upon using other
measures of closeness, e.g. the relative entropy instead of 
(\ref{hellinger}). 

There is yet another quantity that can be employed for evaluating
our approach.  Returning to the discussion above
(\ref{d2}), we generate independently|following the above recipe, and
for a given $\pi_k(x,y)$, $p_k(y)$ and $E_k$|many ($l=1,...,M\gg 1$)
conditional probabilities $\widetilde{p}_k^{[l]}(x|y)$ that hold
$\sum_{xy}p_k(y)\widetilde{p}^{[l]}_k(x|y)E(x,y)=E_k$. Next, we consider
the average: 
\BEA
\label{kima}
\overline{\widetilde{p}}_k(x|y)=\frac{1}{M}\sum_{l=1}^M
\widetilde{p}_k^{[l]}(x|y), 
\EEA
which also corresponds to the known practice of taking averages over
different outcomes of the likelihood maximization.
Eq.~(\ref{kima}) is akin to the Bayesian-average estimator, because for
given observations (for this case $p_k(y)$) it averages over all hidden
parameters consistent with the prior information $E_k$. 

\begin{table}
\begin{center}
\caption{The values of $\Delta D_3$ given by (\ref{d3}) for 
$x= 1, 2, 3$, $y= 1, 2, 3$, $E_1(x,y)=|x-y|$, and $E_2(x,y)=xy$.
The averaging in (\ref{d3}) was taken over $S=300$ samples for $M=10^5$ in (\ref{kima}). We took $\beta=0.95$.
We also presented the analogues of $\Delta D_3$ (denoted by $\Delta K_3$), where
Hellinger's distance in (\ref{d3}) is replaced by the relative entropy:
${\rm dist}[\pi_k,\hat p_k]\to \sum_{xy} \pi_k(x,y)\ln\frac{\pi_k(x,y)}{\hat p_k(x,y)} $.
Both choices support the same conclusion: $\Delta D_3>0$, $\Delta K_3>0$.}
\begin{tabular}{||c|c||}
\hline\hline
  $E_1$   & $\Delta D_3=0.00278$, $\Delta K_3=0.01224$~   \\
\hline
\hline
  $E_2$   & $\Delta D_3=0.00215$, $\Delta K_3=0.01053$~      \\
\hline\hline
\end{tabular}
\end{center}
\label{tab2}
\end{table}

To understand whether $p_k(y)\overline{\widetilde{p}}_k(x|y)$ is a
better estimate of $\pi_k(x,y)$ as compared to $\hat p_k(x,y)$, we look at
averages over independent $\pi_k(x,y)$ [cf.~(\ref{d1}, \ref{d2})]: 
\BEA 
\label{d3}
\Delta D_3=\frac{1}{S}\sum_{k=1}^S \Delta d_k, \qquad
\Delta d_k \equiv {\rm dist}[\pi_k, p_k(y) \overline{\widetilde{p}}_k(x|y)]
- {\rm dist}[\pi_k, \hat p_k(x,y)].
\EEA
Though particular values of $\Delta d_k$ can be negative, the averaged
value $\Delta D_3>0$ is positive showing that $\hat p_k(x,y)$ [given by (\ref{mur1})] is a
better estimate than $p_k(y)\overline{\widetilde{p}}_k(x|y)$; see Table~\ref{tab2}. 

Comparing Table~\ref{tab2} with results of Table~I, we see that
$p_k(y)\overline{\widetilde{p}}_k(x|y)$ is closer to $\pi_k(x,y)$ than a
single random guess $p_k(y){\widetilde{p}}_k(x|y)$|hence the practical
habit of averaging over different outcomes of the maximum-likelihood
method does have a rationale in it|but it is still outperformed by $\hat
p_k(x,y)$. 

\comment{
\begin{figure*}[t] 
\includegraphics[width=7cm]{nonident1.eps} 
\caption{
Hellinger's distance (\ref{hellinger}) between three randomly
generated probabilities $\pi_k(x,y)$ ($k=1,2,3$) and and their estimates
$\hat p_k(x,y)$. To understand the parametric dependence on $\beta$ we
analytically continued (\ref{mur1}) to $\beta>1$ (note that this does 
not anymore correspond to a local maximum of minimum of $L_{\beta>1}$).\\
It is seen that as a function of $\beta$ the distance
is minimized in the vicinity of $\beta=1$, but not necessarily at
$\beta=1$. In fact, the green curve minimizes for $\beta<1$, while other
two curves minimize for $\beta>1$. For all curves we omitted the
vicinity of $\beta=1$ and $\beta>0$, where solutions of (\ref{khr0}, \ref{khr1}) are
unstable (also due to the spurious solution $\delta=1$). This point illustrates
the general tendency of the method to produce unstabilities in the
vicinity of $\beta=1$ and $\beta>1$; cf. the behavior of Hessian 
in this regime. } 
\label{fig1}
\end{figure*}
}

\section{Maximizing the generalized likelihood for $\beta>1$}
\label{beta>1}

We now turn to maximizing $L_{\beta>1}$ over unknown probabilities $\hat
p(x,y)$; cf.~(\ref{gosh}). As seen below, this leads to setting many
unknown probabilities to zero, i.e. making the vector $\{\hat p(x,y)\}$
sparse. Hence the maximization of $L_{\beta>1}$ does not apply to the
problem of solving the observational nonidentifiability, unless this
problem comes with a prior information on the sparsity. Even apart of
such cases, studying ${\rm max}[\, L_{\beta>1}\,]$ is relevant for those
examples, where the maximization of $L_{\beta<1}$ does not provide
sufficiently nontrivial result; see section \ref{margo},
where only the marginal probabilities of $X$ and $Y$ are 
known. As seen below, yet another reason for studying ${\rm max}[\, L_{\beta>1}\,]$ is
that it does have close relations with entropy minimization, a
technique sporadically employed in probabilistic inference
\cite{good_entropy,christensen,watanabe_entropy} (e.g., for the feature
extraction problem \cite{christensen}) and recently discussed in the
context of risk-minimization in decision-making \cite{armen}. 

\comment{
since the structure of ${\rm argmax}[\,L_{\beta>1}]$ is to some extent
universal|i.e. it does not depend on the concrete value of $\beta>1$,
and also on the shape of $L_\beta$; see below for details.  It is also
an interesting computationally complex problem (akin to certain problems
of concave programming) and hence requires approximate methods for its
solution.} 

For simplicity we assume that $N\gg 1$ in (\ref{data}), i.e. 
$\sum_x\hat p(x,y)=p(y)$ holds. 
Hence we use $\hat p(x,y)=\hat p(x|y)p(y)$ and write (\ref{gosh}) as 
\BEA
L_\beta=\frac{1}{\beta}\sum_yp(y)\ln\left[\sum_x \hat p^\beta(x|y)
\right]+\sum_y p(y)\ln p(y), 
\label{udod} 
\EEA 
where $\{\hat p(x|y)\}$ can be taken as
maximization variables. Besides $\sum_x\hat p(x|y)=1$ and $\hat p(x|y)\geq 0$,
there can be additional conditions imposed on the maximization, e.g.
condition (\ref{beri2}). We denote such conditions by ${\cal C}$. Without
such constraints, the maximization of (\ref{udod}) is trivial: since 
$\sum_x\hat p^\beta(x|y)\leq 1$ due to $\beta>1$, the global maximum of
(\ref{udod}) is reached for $\hat p(x|y)=p_{\rm sparse}(x|y)=\delta_{\rm K}(x,x')$, where 
$\delta_{\rm K}(x,x')$ is the Kronecker delta, and where $x'$ is an 
arbitrary value of $X$. Note that the same $p_{\rm sparse}(x|y)$ minimizes 
the entropy 
\BEA
\label{rostok} S_{XY}=
-\sum_{xy}p(y)\hat p(x|y)\ln [p(y)\hat p(x|y)], 
\label{entropy}
\EEA
over
$\{\hat p(x|y)\}$. 
In Appendix \ref{margot} we present a numerical evidence that
the maximizer of (\ref{gosh}) for $\beta>1$ coincides the minimizer of
(\ref{entropy}) under a nontrivial constraint ${\cal C}$ of the known marginal $p(x)$. 

This minimizer corresponds to
the possibly majorizing (i.e. in the sense of majorization \cite{major})
probability vector under constraints ${\cal C}$; see Appendix
\ref{schuro} for details.  To describe it, one introduces
\BEA
\underset{p(x_k|y_l);\, l=1,...,m, k=1,...,n }{ {\rm max}}[\,\hat p(x_k|y_l)p(y_l)\,; {\cal C}].
\label{maxo1}
\EEA
If the maximization in (\ref{maxo1}) is reached at $k=k^*$ and $l=l^*$, 
then the next element of $\{p(x_k|y_l)\}_{l=1,...,m}^{k=1,...,n} $ is found from:
\BEA
\label{maxo2}
\underset{p(x_k|y_l);\, l=1,...,m, k=1,...,n,\, k\not =k^*,\, l\not =l^* }{{\rm max}}[\,\hat p(x_k|y_l)p(y_l)\,; {\cal C}].
\EEA
This process continues|taking at each step all previously found elements as contraints|till 
all elements of $\{p(x|y)\} $ are found. 
Eqs.~(\ref{maxo1}, \ref{maxo2}) emerge as maximizers of a generalized
Schur-convex function; see Appendix \ref{schuro}. We emphasize that
$L_{\beta>1}$ in (\ref{udod}) is not a generalized Schur convex; hence
the relation between the maximizer of $L_{\beta>1}$ and (\ref{maxo1},
\ref{maxo2}) is presently an empiric (numeric) fact that needs further
understanding. 

\comment{
At any rate, we stress that according to (\ref{maxo1}, \ref{maxo2}),
$\{\hat p(x|y)\}$ will normally contain many zeroes. Hence it can be
employed only if there is a prior information on sparsity in $\{\hat
p(x|y)\}$. }

\comment{
\section{To be understood}

\BEA
(p(y_1), p(y_2), p(y_3))=(p_{y\,1}^{\downarrow}\geq p_{y\,2}^{\downarrow}\geq p_{y\,3}^{\downarrow}\,), 
\qquad
(p(x_1), p(x_2), p(x_3))=(p_{x\,1}^{\downarrow}\geq p_{x\,2}^{\downarrow}\geq p_{x\,3}^{\downarrow}\,).
\EEA

$\bullet$ Derivation of $L_\beta$ {\it a la} Shore-Johnson.

$\bullet$ Various versions of likelihood (taken from recent arxiv by
Vikesh Siddhu): hedged maximum likelihood, maximum likelihood-maximum
entropy estimator. 

Pseudolikelihood is discussed in: 
Inverse statistical problems: from the inverse Ising problem to data science
H. Chau Nguyen, R. Zecchina, J. Berg, arxiv. 

$\bullet$ Look up papers in Cognition/Decision Utility/Free energy. Some of them are potentially useful. 
}

\section{Summary and open problems}

How to solve nonidentifiability in parameter determination of mixiture
models? We proposed an answer that applies to observational
nonidentifiability, where the full model (including hidden variables) is
identifiable, but the observed (marginal) model is not; see section
\ref{defo}.  Marginalizing decreases the information available about the
unknown parameter(s) \cite{cox}. This general point can be illustrated by the
behavior of the Fisher information that decreases upon marginalizing
\cite{cox}. Here we focus on the extreme case, when the information
about the parameter is lost completely. This is the phenomenon of
observational nonidentifiability, where the maxima of the marginal
likelihood are (infinitely) degenerate. In contrast to most general
instances of nonidentifiability (which e.g. can follow from a trivial
overparametrization), this particular form is not hopeless to solve,
precisely because the full model (including the unobserved or hidden
variables) is identifiable. 

The presented method amounts to generalizing the marginal likelihood function via
$L_\beta(\theta)$, where $\theta$ is the unknown parameter(s), and
$\beta>0$ is an analogue of inverse temperature from statistical
mechanics; see section \ref{gena}. For $\beta=1$ we recover the usual
marginal likelihood, while $L_\infty(\theta)$ amounts to the
h-likelihood, where the value of hidden variables is replaced by its MAP
(maximum aposteriori) estimate. $L_\beta(\theta)$ is constructed by
analogy to the statistical physical free energy, where $\beta$ plays the
role of inverse temperature; see section \ref{gena}. The generalization
is motivated by the fact that $L_\beta(\theta)$ inherits some of useful
features of $L_1(\theta)$; see section \ref{gena}. 

Maximizing $L_\beta(\theta)$ instead of $L_1(\theta)$ can lead to
reasonable predictions if the value of $\beta$ is chosen correctly. We
treated several models and argued that the optimal value of $\beta$ is
close to, but (strictly) smaller than one. In particular, results
predicted by $L_\beta(\theta)$ are better than those obtained via what
one can call a practitioner's attitude towards nonidentifiability, i.e.
picking up a random maximum of $L_1$, or averaging over many such
(randomly selected) maxima; see section \ref{compo}. The check was
carried out numerically by assuming that the initial data is distributed
randomly in a sufficiently unbiased way. We have shown that maximizing
$L_{\beta\lesssim 1}(\theta)$ relates to the maximum entropy method; see
section \ref{uraa}. Likewise, the maximization of $L_{\beta> 1}(\theta)$
relates with minimizing the entropy; see section \ref{beta>1}. There are
also some analogies between $L_\beta(\theta)$ and conditional Renyi
entropies \cite{renyi1,renyi2}. 

Several pertinent questions are left open and should motivate further
research. 
{\it (i)} Results and methods of section \ref{compo}|that compares predictions of
$L_\beta(\theta)$ with random selections|should be studied
systematically on an analytical base. 
{\it (ii)} How $L_\beta(\theta)$ applies to
effective nonidentifiability?  
{\it (iii)} Asymptotic analysis of $L_\beta(\theta)$ that should
link it to a (generalized?) Fisher information. 
{\it (iv)} The relation between maximizing $L_{\beta>1}(\theta)$ and entropy minimization should be
clarified; see section \ref{beta>1}. So far it is restricted to a
perturbation argument (see section \ref{uraa}) and numerical examples;
cf. Appendix \ref{margot}.  
{\it (v)} How $L_\beta(\theta)$ applies
to image restoration problems that also frequently suffer from
observational nonidentifiability issues \cite{image}?


\comment{What did we learn generally and from the model:

Three main themes: (1) $L_\beta$ is a simple generalization of the usual
likeliihood $L_1$ that inherits features of $L_1$.  It is suitable for
resolving the nonidentifiability in the sense of (\ref{nono},
\ref{star}), i.e. when the nonidentifiability comes due to partial
observability. (2) Maximization of $L_{\beta<1}$ in the model with
unknown probabilities relates to the entropy maximization. The optimal
value of $\beta$ locates in the vicinity of $1$ (but $\beta=1$ is
excluded). (3) In contrast to $L_{\beta<1}$, we cannot apply
$L_{\beta<1}$ if the data on the observed variable is incomplete.
Maximization of $L_{\beta>1}$ relates to the entropy minimization. Here
simple constraints lead to intractable maximizers, whose explicit form
can be obtained only via enumeration over a large number of variables.
However, sub-optimal solutions can be simple, as we demonstrated. Such
maximizers have a possibly large of number of zeroes, i.e. they are
applicable if there is a prior information on sparsity. 

Perspective: }

\acknowledgements

It is a pleasure to acknowledge many useful discussions with Narek Martirosyan.
I thank Aram Galstyan for support and discussions and Gevorg Karyan for a useful remark. 

This research was supported by ISTC Joint Research Grant Program
{\it Parameter learning in nonidentifiable models}, and 
by SCS of Armenia, grants No. 18RF-015 and No. 18T-1C090.

\appendix 

\section{Two alternative approaches}

\subsection{Bayesian approach to observational nonidentifiability}
\label{bayo}

Here we outline a Bayesian approach to the observationally
nonidentifiable model discussed in section \ref{simpo}. First of all,
recall from (\ref{wes}) that $p_{gh}(x,y)=p(x,y|g,h)$ is a conditional
probability.  Given an observation $Y=y$ we want to exclude the
parameter $h$ so as to gain information on the parameter $g$ via a
conditinal probability $p(g|y)$. To this end, we have to come up with
prior probabilities of $h$ and $g$. We make the simplest assumption that
a priori they are independent: $p(h,g)=p(h)p(g)$ and that the prior
$p(h)$ is noninformative. In the Bayesian approach this means that we
have to take \cite{jaynes} (depending on the possible range of $h$):
\BEA
\label{shle}
p(h)\propto 1 \qquad {\rm if}\qquad \infty>h>-\infty,\\
p(h)\propto 1/h \qquad {\rm if}\qquad \infty>h>0.
\label{mle}
\EEA
Note that both priors probability densities (\ref{shle}, \ref{mle}) are improper, i.e. they are
not normalizable. Improper priors can still lead to useful applications
\cite{jaynes}. 

The first step in calculating $p(g|y)$ is to study $p(x,y|g)$ from (\ref{wes}) and (\ref{shle})
\BEA
p(x,y|g)\overset{?}{\propto} \int \d h\, p(x,y|g,h) p(h).
\label{budo}
\EEA
Now (\ref{shle}, \ref{mle}) and (\ref{wes}) show that the integral in the right-hand-side of 
(\ref{budo}) does not exist, i.e. the Bayesian approach with non-informative priors
is blocked already at its first step. If a proper prior is available instead of (\ref{shle}, \ref{mle}),
then the Bayesian approach does work. We do not dwell into this, 
since we assume that no prior information is available.

\subsection{Decision theory approach: attempts to build a maximin
estimator for an observationally non-identifiable model }
\label{maxmayo}

Following the basic tenets of the decision theory approach \cite{cox},
we shall attempt to build a maximin estimator for the model model
discussed in section \ref{simpo}.  The virtue of such an estimator is
that its construction does not need prior probabilities for unknown
parameters \cite{cox}. 

Starting from (\ref{wes}, \ref{pretoria}) and assuming that $y$ is observed we construct
\BEA
\label{omm}
{\rm dist}[p_{\hat g(y)\, \hat{h} (y)}(x|y), p_{g h}(x|y)  ]=1-\sum_x
\sqrt{p_{\hat g(y)\, \hat{h} (y)}(x|y) p_{g h}(x|y)}\\
=1-\frac{\cosh\left[ \frac{g+\hat g(y)}{2}+\frac{y(h+\hat h(y))}{2}  \right]}
{\sqrt{\cosh\left[ g+yh \right]\,\cosh\left[ \hat g(y)+y\hat h(y) \right]}},
\label{ki}
\EEA
where the choose to work with Hellinger's distance, and where $\hat g(y)$ and $\hat{h} (y)$ are estimators of (resp.) $g$
and $h$ given the observation $y$. Together with (\ref{omm}), one also employs the distance, which is averaged over
observations \cite{cox}:
\BEA
\label{ob}
&& \sum_{y}p_{gh}(y)\, {\rm dist}[p_{\hat g(y)\, \hat{h} (y)}(x|y), p_{g h}(x|y)  ]\\
&& =1-\frac{\cosh[u]}{\cosh[u]+\cosh[v]}\,\frac{\cosh[(u+\widetilde{u})/2]}{\sqrt{\cosh[u]\cosh[\widetilde u]}}
-\frac{\cosh[v]}{\cosh[u]+\cosh[v]}\,\frac{\cosh[(v+\widetilde{v})/2]}{\sqrt{\cosh[v]\cosh[\widetilde v]}},
\label{obe}
\EEA
where in (\ref{obe}) we defined
\BEA
u\equiv h+g>0, \qquad v\equiv h-g, \qquad 
\widetilde u\equiv \hat h(1)+\hat g(1)>0, \qquad \widetilde v\equiv \hat h(-1)-\hat g(-1).
\label{ula}
\EEA
Note that the constraints $u>0$ and $\widetilde u>0$ come from our assumption on $g>0$ and $h>0$.

The maximin estimator takes the worst case (i.e. the maximal distance)
with respect to unknown parameters $g$ and $h$, and then minimizes this
worst case over the estimators $\hat h(y)$ and $\hat g(y)$ \cite{cox}.
In principle, this procedure can be applied to either (\ref{ki}) or
(\ref{obe}). We shall start by applying it to (\ref{ki}). We note from
(\ref{ki}, \ref{ula}):
\BEA
&& 1- {\rm dist}[p_{\hat g(y)\, \hat{h} (y)}(x|1), p_{g h}(x|1)  ]=
\frac{\cosh[(u+\widetilde{u})/2]}{\sqrt{\cosh[u]\cosh[\widetilde u]}},\\
&& 1- {\rm dist}[p_{\hat g(y)\, \hat{h} (y)}(x|-1), p_{g h}(x|-1)  ]=
\frac{\cosh[(v+\widetilde{v})/2]}{\sqrt{\cosh[v]\cosh[\widetilde v]}}.
\EEA
The first step amounts to maximizing the distance over unknown $g$ and $h$, i.e. over $u>0$ and $v$:
\BEA
\label{g21}
{\rm min}_{u>0}\, \frac{\cosh[(u+\widetilde{u})/2]}{\sqrt{\cosh[u]\cosh[\widetilde u]}}
                &=& \frac{e^{\widetilde u/2}}{\sqrt{2\cosh \widetilde u}} \quad {\rm for}\quad \widetilde u<-\ln[\sqrt{2}-1],\\
\label{g22}
                &=& \frac{\cosh[\widetilde u/2]}{\sqrt{\cosh \widetilde u}} \quad {\rm for}\quad \widetilde u>-\ln[\sqrt{2}-1],\\
\label{g23}
{\rm min}_{v}\,\frac{\cosh[(v+\widetilde{v})/2]}{\sqrt{\cosh[v]\cosh[\widetilde v]}}
                &=& \frac{e^{-|\widetilde v|/2}}{\sqrt{2\cosh \widetilde v}},
\EEA
where the last relation is deduced for $v\to\pm\infty$ depending on the sign of $\widetilde v$.

At the second step we should minimize the distance over estimators, i.e. 
(\ref{g21}, \ref{g22}) is to be maximized over $\widetilde u$, while 
(\ref{g23}) is to be maximized over $\widetilde v$. This step is supposed
to define those estimators. We get from (\ref{g21}--\ref{g23}):
\BEA
\label{bru}
&& \widetilde u=0,\infty, \\ 
&& \widetilde v=0,
\label{gru}
\EEA
where the two values $0$ or $\infty$ for $\widetilde u$ in (\ref{bru}) come from (resp.) 
(\ref{g21}) and (\ref{g22}). While $\widetilde v=0$ in (\ref{gru}) seems reasonable (though incomplete) value for the estimator, neither of 
$\widetilde u=0$ or $\widetilde u=\infty$ is meaningful. Hence the maximin strategy applied to 
(\ref{ki}) does not leas to sensible estimators. 

When applying the strategy to (\ref{obe}), we note that (\ref{g23})$\leq$(\ref{g21}) and (\ref{g23})$\leq$(\ref{g22}) for
all allowed values of $\widetilde u$ and $\widetilde v$. Hence we find
\BEA
{\rm min}_{u>0,\, v}\left[
\frac{\cosh[u]}{\cosh[u]+\cosh[v]}\,\frac{\cosh[(u+\widetilde{u})/2]}{\sqrt{\cosh[u]\cosh[\widetilde u]}}
-\frac{\cosh[v]}{\cosh[u]+\cosh[v]}\,\frac{\cosh[(v+\widetilde{v})/2]}{\sqrt{\cosh[v]\cosh[\widetilde v]}}
\right]=\frac{e^{-|\widetilde v|/2}}{\sqrt{2\cosh \widetilde v}},
\label{dru}
\EEA
where the last relation is again deduced for $v\to\pm\infty$. The maximization of (\ref{dru}) over $\widetilde v$
brings us sback to (\ref{gru}). Again nothing reasonable is produced for $\widetilde u$: the maximin method does not work for this example.

\section{Maximization of $L_\beta$ under two constraints}
\label{two}

Consider the maximization of $L_\beta$ given by (\ref{gosh}) under two constraints
[cf.~section \ref{oslo}]
\BEA
\label{beri}
&& \sum_x \hat p(x,y)=p(y),\qquad \sum_{xy}E(x,y) \hat p(x,y)=E, 
\label{tari}
\EEA
where $E(x,y)$ is a function of $x$ and $y$ with a known average $E$.
The Lagrange function reads:
\BEA
{\cal L}_\beta=
\frac{1}{\beta}\sum_yp(y)\ln\left[\sum_x \hat p^\beta(x,y) \right]
- \sum_{xy}\delta(y)\hat p(x,y)
-\gamma \sum_{xy}E(x,y)\hat p(x,y),
\EEA
where $\delta(y)$ and $\gamma$ refer to (resp.) (\ref{beri}) and (\ref{tari}).
Now $\frac{\partial {\cal L}_\beta }{\partial\hat p(x,y) }=0$ leads to 
\BEA
\label{mur}
\frac{p(y) \hat p^{\beta-1}(x,y)   }{\sum_{\bar{x}} \hat p^\beta(\bar x,y) }=\delta(y)+\gamma E(x,y),
\EEA
which is solved as 
\BEA
\label{surok}
\hat p(x,y)=p(y)\,\frac{\left[\delta(y)+\gamma E(x,y)   \right]^{\frac{1}{\beta-1}}}
{\sum_{\bar{x}} \left[\delta(y)+\gamma E(\bar x,y)   \right]^{\frac{\beta}{\beta-1}}  }.
\EEA
Here $\delta(y)$ and $\gamma$ are found from (resp.) (\ref{beri}) and (\ref{tari}). Eventually, $\gamma$ can be expressed
via $\delta(y)$, which is found from (\ref{khr}):
\BEA
&& \gamma =\frac{1}{E}\left( 1-\sum_yp(y)\delta(y) \right),\\
&& \sum_x {\left[\delta(y)+\gamma E(x,y)   \right]^{\frac{1}{\beta-1}}} =
{\sum_{{x}} \left[\delta(y)+\gamma E( x,y)   \right]^{\frac{\beta}{\beta-1}}  }.
\label{khr}
\EEA

\clearpage

\section{The relevance of various constraints in the maximum entropy method}
\label{tra}

The maximum entropy method addresses the problem of recovering unknown
probabilities $\{q(z_k)\}_{k=1}^n$ of a random variable
$Z=(z_1,...,z_n)$ via maximization of the entropy 
\BEA
\label{ento}
S[q]= -\sum_{k=1}^n
q(z_k)\ln q(z_k). 
\EEA
subject to certain constraints on $q$ and $Z$
\cite{jaynes,jaynes_2,skyrms,enk,cheeseman}. These constraints are
assumed to come as a prior information, within its standard formulation
the method does not determine the type and a number of those
constraints; the only (obvious) requirement from the method is that the
result of maximization is unique. The intuitive rationale of the method
is that provides the most unbiased choice of probability compatible with
the constraints. 

One way of recovering the constraints is to look at (necessarily noisy)
data.  If this way is followed in detail, it can give some
recommendations on selecting the constraints, or at least on determining
their relative relevance.  Below we shall present some preliminary
results to this effect within. Since the results are preliminary, we
shall not attempt to generalize them towards the likelihood
$L_{\beta<1}$. 

A standard way of checking an inference method is to assume that the
true probabilities are known. Hence  we shall start by assuming that
we know the probabilities $\{q(z_k)\}_{k=1}^n$ of $Z=(z_1,...,z_n)$. From 
$\{q(z_k)\}_{k=1}^n$ we generate a finite i.i.d. sample 
\BEA
\label{sampo}
{\cal S}_M=(z_{i_1},...,z_{i_M})
\EEA
of length $M$. Various constraints are now to be recovered from (\ref{sampo}).
Here are several examples

-- We can apply no constraint at all and just maximize the entropy:
\BEA
q^{[0]}(z_k)=\frac{1}{n}.
\label{grundig}
\EEA

-- After calculating the empiric mean of (\ref{sampo}),
\BEA
\label{first}
\mu=\frac{1}{M}\sum_{u=1}^M z_{i_u},
\EEA
we can take it as an estimate for the true average $\sum_{k=1}^n q(z_k)z_k$, and recover approximate probabilities
$\{q^{[1]}(z_k)\}_{k=1}^n$ via maximizing (\ref{ento}) subject to a constraint: $\sum_{k=1}^n q^{[1]}(z_k)z_k=\mu$.
It is well-known \cite{jaynes,jaynes_2} that this maximization leads to 
\BEA
\label{s1}
q^{[1]}(z_k)=\frac{e^{-\beta z_k}}{\sum_l e^{-\beta z_l} },
\EEA
where $\beta$ is determined from $\sum_{k=1}^n q^{[1]}(z_k)z_k=\mu$.

-- The empiric means is certainly not the only information contained in the 
sample; e.g. one can estimate as well the second moment:
\BEA
\label{second}
\mu_2=\frac{1}{M}\sum_{u=1}^M z^2_{i_u},
\EEA
and maximize (\ref{ento}) under two contraints (\ref{first}) and (\ref{second}):
\BEA
\label{s2}
q^{[1+2]}(z_k)=\frac{e^{-\beta_1 z_k-\beta_2 z_k^2}}{\sum_l e^{-\beta_1 z_l-\beta_2 z_l^2} },
\EEA
where $\beta_1$ and $\beta_2$ are determined from $\sum_{k=1}^n q^{[1+2]}(z_k)z_k=\mu$ and
$\sum_{k=1}^n q^{[1+2]}(z_k)z_k^2=\mu_2$.

-- It is the standard lore of statistics that for relatively short samples, the empiric median is 
a better (more robust) estimator than the empiric mean. Thus we should pay attention to the 
median as a constraint in the entropy maximization. Recalling the definition of the median
${\rm Md}$ for given (discrete-variable) probabilities, the maximum of (\ref{ento}) under a fixed 
median is made obvious with the following example for $n=4$ (assuming for simplicity that $z_1<z_2<z_3<z_4$):
\BEA
\label{y1}
{\rm Md}=z_1: \quad {\rm argmax}_q[\,S[q]\, ]=\frac{1}{2}\left(1+\epsilon,\frac{1-\epsilon}{3}, \frac{1}{3}, \frac{1}{3}\right),\\
{\rm Md}=z_2: \quad {\rm argmax}_q[\,S[q]\, ]=\frac{1}{2}\left(\frac{1}{2},\frac{1+\epsilon}{2}, \frac{1-\epsilon}{2}, \frac{1}{2}
\right),\\
{\rm Md}=z_3: \quad {\rm argmax}_q[\,S[q]\, ]=\frac{1}{2}\left(\frac{1}{3},\frac{1}{3}, \frac{1+\epsilon}{3}, 1-\epsilon\right),\\
{\rm Md}=z_4: \quad {\rm argmax}_q[\,S[q]\, ]=\frac{1}{2}\left(\frac{1}{3},\frac{1}{3}, \frac{1-\epsilon}{3}, 1+\epsilon\right),
\label{y2}
\EEA
where $\epsilon>0$ is infinitely small. We kept it for confirming that the median is indeed equal to its fixed value, but $\epsilon$
can be neglected in actual calculations. Eqs.~(\ref{y1}--\ref{y2}) easily generalize to an arbitrary finite $n$. 

Now the median ${\rm Md}$ will estimated from the finite sample (as an empiric median), and the maximum entropy probabilities
recovered according to (\ref{y1}--\ref{y2}) will be denoted as $\{q^{\rm [md]}(z_k)\}_{k=1}^n$

We can calculate how close are the above estimates from the true probabilities $q=\{q(z_k)\}_{k=1}^n$:
\BEA
d_0={\rm dist}[q^{[0]},q], \quad
d_1={\rm dist}[q^{[1]},q],\quad 
d_{1+2}={\rm dist}[q^{[1+2]},q],\quad
d_{\rm md}={\rm dist}[q^{[{\rm md}]},q],
\label{we}
\EEA
where ${\rm dist}[.,.]$ can be e.g. the Hellinger distance: 
\BEA
{\rm dist}[q^{[0]},q]\equiv 1-\sum_{k=1}^n\sqrt{ q(z_k)\, q^{[0]}(z_k)}.
\label{hellinger2}
\EEA

Besides $d_0$, quantities defined in (\ref{we}) 
are random variables together with the sample (\ref{sampo}). Hence we shall average them over 
${\cal M}\gg 1$ independently generated samples, keeping the sample length $M$ fixed. The averaged quantities 
will be denoted as 
\BEA
\label{korund}
\langle d_1\rangle, \quad \langle d_{1+2}\rangle, \quad \langle d_{\rm md}\rangle.
\EEA
Together with $d_0$ they depend on $q=\{q(z_k)\}_{k=1}^n$. Besides
quantities in (\ref{korund}) we shall also study their averages over
$q=\{q(z_k)\}_{k=1}^n$: we generate randomly ${\cal N}$ probabilities
(the mechanism for this is discussed in section \ref{compo} of the main
text), and average $\langle d_1\rangle$, $\langle d_{1+2}\rangle$,
$\langle d_{\rm md}\rangle$, and $d_0$ over them.  The results will be
denoted by
\BEA
\label{bat}
\overline{d^{}}_1, \quad \overline{d^{}}_{1+2}, \quad \overline{d^{}}_{\rm md}, \quad \overline{d^{}}_0. 
\EEA

Table~\ref{ta1} presents a numerical illustration for quantities defined
in (\ref{we}--\ref{bat}). It is seen that when $M$ is larger, but
comparable to $n$ ($M=7$ and $n=6$ in Table~\ref{ta1}), the situation is
so noisy that samples do not provide information from the viewpoint of
the constraints studied \footnote{This does not mean that short samples 
provide no information whatsoever. This means that the proper information 
extraction mechanism from such samples is yet to be 
found.}. This means that the no-constraint solution
(\ref{grundig}) is always better because in the majority of cases we get
${\rm min}[\langle d_1\rangle, \langle d_{1+2}\rangle, \langle d_{\rm
md}\rangle, d_0]=d_0$, and because ${\rm min}[\overline{d^{}}_1, 
\overline{d^{}}_{1+2}, \overline{d^{}}_{\rm md}, 
\overline{d^{}}_0]=\overline{d^{}}_0$. For such values of $M$ employing the 
above constrained solutions will just amount to an overfitting (of noise).

For a larger $M$ ($M=11$ and $n=6$ in Table~\ref{ta1}), we see that (\ref{s1}) is the best constraint
in one sense, since now the solution (\ref{s1}) provides
a smaller average distance from the true solution: ${\rm
min}[\overline{d^{}}_1, \overline{d^{}}_{1+2}, \overline{d^{}}_{\rm md},
\overline{d^{}}_0]=\overline{d^{}}_1$. However, in the second sense 
(\ref{grundig}) is still better, because the percentage of cases, where
${\rm min}[\langle
d_1\rangle, \langle d_{1+2}\rangle, \langle d_{\rm md}\rangle,
d_0]=d_0$ is still the largest one.
Applying the median solution or
the second-order solution (\ref{s2}) lead to worse results.
Table~\ref{ta1} shows that the solution based on the median is always
worse than some of the other solutions. 

The second-order solution (\ref{s2}) becomes the best solution for
$M\geq 21$; see Table~\ref{ta1}. This holds in terms of the average
distance: ${\rm min}[\overline{d^{}}_1, \overline{d^{}}_{1+2},
\overline{d^{}}_{\rm md}, \overline{d^{}}_0]=\overline{d^{}}_{1+2}$, and
also in terms of the percentage of cases, where ${\rm min}[\langle
d_1\rangle, \langle d_{1+2}\rangle, \langle d_{\rm md}\rangle,
d_0]=d_{1+2}$. Increasing $M$ more just confirms this trend, i.e.
improves the quality of (\ref{s2}) in both senses. Interestingly, the
percentage of cases, where ${\rm min}[\langle d_1\rangle, \langle
d_{1+2}\rangle, \langle d_{\rm md}\rangle, d_0]=d_{1}$ is relatively
stable for larger values of $M$: in more than $1/6$ of cases the
first-order solution (\ref{s1}) is still better than other solutions,
even for $M=101$; see Table~\ref{ta1}. 

Our (preliminary) conclusions are summarized as follows: {\it (i)}
The median is not a relevant constraint for the maximum entropy
method. It is never better than the average. {\it (ii)} The latter
solution does overfit for short samples ($M\simeq n$), where having
no constraints whatsoever is better than fixing the average. {\it (iii)}
For sufficiently long samples fixing the first and second moments 
outperforms other solutions, but the average constraint does stay
reasonable even for larger sample lengths.

\begin{table}[h!]
\centering
\tabcolsep0.038in \arrayrulewidth0.3pt
\renewcommand{\arraystretch}{0.9}
\begin{tabular}{l|cccc||cccc}
\hline\vspace{0.1cm}
$M$ & $\%\, {\rm min}=\langle d_1\rangle$ & $\%\, {\rm min}=\langle d_{1+2}\rangle$ & $\%\, {\rm min}=\langle d_{\rm med}\rangle$ 
& $\%\, {\rm min}=d_0$ & $\overline{d^{}}_1$ & $\overline{d^{}}_{1+2}$ & $\overline{d^{}}_{\rm md}$ & $\overline{d^{}}_0$ \\
\hline
7  & 18 & 4  & 8  & 70 & 0.06194 & 0.07713 & 0.06771 & \underline{0.05535} \\ 
11 & 25 & 18 & 7  & 50 & \underline{0.05548} & 0.05829 & 0.06212 & 0.05656 \\ 
21 & 24 & 41 & 2  & 33 & 0.04731 & \underline{0.04421} & 0.05894 & 0.05350 \\ \hline
31 & 27 & 45 & 1  & 27 & 0.04583 & \underline{0.04125} & 0.05935 & 0.05520 \\ 
41 & 29 & 51 & 2  & 18 & 0.05091 & \underline{0.04302} & 0.06311 & 0.05970 \\ 
61 & 24 & 64 & 2  & 10 & 0.04628 & \underline{0.03531} & 0.05459 & 0.05430 \\ 
101& 18 & 71 & 0  & 11 & 0.04296 & \underline{0.03567} & 0.05519 & 0.05179 \\ \hline
\end{tabular}
\caption{ For a set-up of a dice: $n=6$ and $z_k=k$ ($k=1,...,6$) we show various quantities defined above and below (\ref{we}).
The ${\rm dist}[.,.]$ in (\ref{we}) was chosen to be the Hellinger distance. $M$ is the number of samples. The average
in (\ref{korund}) is defined over $10^4$ indepedent samples generated via the method described in section \ref{compo}. The average
in (\ref{bat}) is defined over $100$ realizations of probabilities. 
Now $\%\, {\rm min}=\langle d_1\rangle$ means the percentage of the relation
$\%\, {\rm min}[\langle d_1\rangle, \langle d_{1+2}\rangle, \langle d_{\rm md}\rangle, d_0]=d_1$
among those $100$ cases; e.g. $\%\, {\rm min}=\langle d_1\rangle\to 18$ means that in 18 cases out of 100 we
got ${\rm min}[\langle d_1\rangle, \langle d_{1+2}\rangle, \langle d_{\rm md}\rangle, d_0]=\langle d_1\rangle$. The minimal among 
$\overline{d^{}}_1$, $\overline{d^{}}_{1+2}$, $\overline{d^{}}_{\rm md}$, and $\overline{d^{}}_0$ 
is underlined. 
}
\label{ta1} 
\end{table}

\clearpage

\section{Generalized Schur-convexity}
\label{schuro}

We shall briefly review implications of the generalized Schur-convexity
\cite{major} for maximizing functions similar to (\ref{udod}). Though we
were not able to show that (\ref{udod}) is generalized Schur-convex,
numerical results show the Schur-convex maximizers provide a good
description of local maxima for (\ref{udod}). 

Let we are given a differentiable function $\Phi(x; u)$ of two vectors:
$x=(x_1,...,x_M)$ and $u=(u_1,...,u_M)$. Both vary on compact subsets of
$\mathfrak{R}^M$, and $u_k\not=0$ for all $k=1,...,M$. Let us assume
that $\Phi(x; u)$ is Schur-convex \cite{major}:
\BEA
\label{amba2}
(x_k-x_l)
\left[\frac{1}{u_k}\frac{\partial \Phi}{\partial x_k}-\frac{1}{u_{l}}
\frac{\partial \Phi}{\partial x_{l}}\right]\geq 0, \qquad k,l=1,..,M-1.
\EEA
Let ${\cal D}$ be the set of vectors that are ordered as: $x_1\geq ...\geq x_M$.
Denote 
\BEA
\label{gluki}
z_\ell\equiv\sum_{k=1}^\ell u_kx_k, \quad \ell=1,...,M,
\EEA
and note that $\Phi(x; u)$ can be written as
\BEA
\label{oni}
\Phi(x;u)=\Phi(\, \frac{z_1}{u_1},  \frac{z_2-z_1}{u_2},...,  \frac{z_M-z_{M-1} }{u_M} \,;u)\equiv \widetilde{\Phi}(z_1,...,z_{M}).
\EEA
Now for $x\in {\cal D}$, $\widetilde{\Phi}(z_1,...,z_{M})$ is a
non-decreasing function of $z_1$, ... $z_{M-1}$, because then (\ref{amba2})
reduces to 
\BEA
\label{amba}
\frac{1}{u_k}\frac{\partial \Phi}{\partial x_k}-\frac{1}{u_{k+1}}
\frac{\partial \Phi}{\partial x_{k+1}}\geq 0, \qquad k=1,..,M-1,
\EEA
and then (\ref{amba}) implies 
\BEA
\label{kao}
\frac{\partial \widetilde{\Phi} }{\partial z_k}\geq 0 \quad {\rm for}\quad
k=1,...,M-1 \quad {\rm and}\quad x\in {\cal D}. 
\EEA
Let us denote by ${\cal A}$ the set of all vectors $x$ that hold
\BEA
\sum_{k=1}^M u_kx_k=1.
\label{brutos}
\EEA
Eq.~(\ref{kao}) will show how to maximize $\Phi(x;u)$ over $x\in{\cal A}\cap{\cal D}$. 
First consider two vectors, $x\in {\cal D}\cap{\cal A}$ and $y\in {\cal D}\cap{\cal A}$. Eq.~(\ref{gluki}) and
conditions (\ref{kao}) imply that if:
\BEA
\label{glu}
\sum_{k=1}^\ell u_kx_k\geq \sum_{k=1}^\ell u_ky_k, \quad \ell=1,...,M-1,
\EEA
then
\BEA
\label{on}
\Phi(x;u)\geq \Phi(y;u).
\EEA
Eqs.~(\ref{amba2}, \ref{glu}) refer to the concept of $u$-majorization, while $\Phi(x;u)$ 
in (\ref{on}) is a $u$-Schur-convex function \cite{major}.

Now ${\rm argmax}_{x\in{\cal A}\cap{\cal D}}[\, \Phi(x,u)
\,]$ is found as follows: one first finds ${\rm max}_{x_1\in{\cal
A}\cap{\cal D}}[\, u_1x_1 \,]$. Then taking this maximized value as a
condition one obtains ${\rm max}_{x_2\in{\cal A}\cap{\cal D}}[\, u_2x_2
\,]$, then under two previous conditions one finds ${\rm
max}_{x_3\in{\cal A}\cap{\cal D}}[\, u_3x_3 \,]$ {\it etc}. 

We generalize the above reasoning taking instead of ${\cal D}$
any other ordering: $x\in{\cal D}^{\pi}$ means that $x_{\pi_1}\geq ...\geq x_{\pi_M}$,
where $\pi$ is a certain permutation of indices $1,...,M$. 
Conditions (\ref{amba2}, \ref{brutos}) stay without changes. 

To obtain $x^*\equiv{\rm argmax}_{x\in{\cal A}}[\, \Phi(x,u) \,]$ under
(\ref{amba2}) and (\ref{brutos}) (i.e. without imposing any condition
$x\in{\cal D}^{\pi}$ for a specific $\pi$), we shall optimize the above
construction over all possible ${\cal D}^\pi$. Thus one first finds
\BEA
\label{0}
\underset{1\leq k\leq M}{\rm max}\,\underset{x\in{\cal A}}{\rm max}[\, u_kx_k\,].
\EEA
If this maximum is reached at a certain value $x_{j_1}^*$ of $x_{j_1}$, then one looks at 
\BEA
\label{00}
\underset{1\leq k\leq M,k\not=j_1}{\rm max}\, \underset{x\in{\cal A}}{\rm max}[\, u_kx_k\,].
\EEA
If the maximum in (\ref{00}) is reached at a certain value $x_{j_2}^*$ of $x_{j_2}$, 
then the next maximization excludes both $j_1$ and $j_2$; and so on till all elements
of $x^*\in{\cal A}$ will be found.

Returning to the problem stated by the maximization of (\ref{udod}), we
note that the index $k$ in (\ref{brutos}, \ref{amba}) corresponds to the
double index $(x,y)$, where $M=n m$, while $x$ and $u$ refer to $\{\hat
p(x|y)\}_{x,y}$ $\{p(y)\}_{x,y}$, respectively.  Eq.~(\ref{brutos}) then
holds due to normalization. Likewise, ${\cal A}$ is defined from
relevant constraints, e.g. from (\ref{op}). But conditions (\ref{amba2})
for $L_{\beta>1}$ do not hold, since the left-hand-side of (\ref{amba2})
amounts to
\BEA
\left[\hat p(x|y)-\hat p(x'|y')\right]
\left(\frac{\hat p^{\beta-1}(x|y)}{\sum_{\bar{x}} \hat p^\beta(\bar{x}|y)  }
-\frac{\hat p^{\beta-1}(x'|y')}{\sum_{\bar{x}} \hat p^\beta(\bar{x}|y')  }\right),
\EEA
which is generally not nonnegative.
In contrast, the negative average entropy: $\sum_y p(y)\sum_x \hat p(x|y)\ln \hat p(x|y)$
does hold (\ref{amba2}):
\BEA
\left[\hat p(x|y)-\hat p(x'|y')\right] \left[\ln\hat p(x|y)-\ln\hat p(x'|y')\right]\geq 0.
\EEA

\section{Maximization of $L_{\beta>1}$ 
for known marginals of $X$ and $Y$ illustrated via examples}
\label{margot}

There are infinitely many joint probabilities $\hat p(x,y)$ with given
marginals $p(y)$ and $p(x)$
\cite{copulas,cohen_zap,finch}. One can ask about the
simplest joint probabilities compatible with given marginals
\cite{good,kullback}. Such a probability can be employed as a
null-hypothesis and serve as a starting point for further
approximations. It is well-known that the maximal entropy reasoning
leads to the factorized joint probability $\hat p(x,y)=p(x)p(y)$
\cite{good}, which we also got from maximizing $L_{\beta<1}$; see
section \ref{margo}. Below we show numerically that the maximization of
$L_{\beta>1}$ leads to a different and unique prediction for $\hat
p(x,y)$ that agrees with (\ref{maxo1}, \ref{maxo2}). Hence it agrees
with minimizing the joint entropy (\ref{rostok}) under the constraint of
given marginals. This is a well-known problem, because (for given
marginals) it is equivalent to maximizing the mutual information between
$X$ and $Y$; see
\cite{min_entropy_marg_1,min_entropy_marg_2,min_entropy_marg_3} for
recent discussions.

Let us assume that both $X$ and $Y$ assume 3 values $(x_1, x_2, x_3)$
and $(y_1, y_2, y_3)$, respectively.  Here is an example for the
(global) maximizer of (\ref{udod}) that we presented in the form of
(\ref{op}) with numeric values of $\hat p(x|y)$ written in bold:
\BEA
\label{xu1}
&& (\, p(y_1), p(y_2), p(y_3)\,)=(0.1, 0.3, 0.6),  \\
\label{xu2}
&& p(x_1)=0.55= {\bf 0}\times 0.1+  {\bf 0}\times 0.3+ {\bf \frac{55}{60}}\times 0.6, \\
&& p(x_2)=0.25= {\bf 0}\times 0.1+ {\bf \frac{25}{30}}\times 0.3+ {\bf 0}\times 0.6,  \\
&& p(x_3)=0.20= {\bf 1}\times 0.1+ {\bf \frac{5}{30}}\times 0.3+ {\bf \frac{5}{60}}\times 0.6.
\label{xu4}
\EEA
Eqs.~(\ref{xu2}--\ref{xu4}) follow (\ref{maxo1}, \ref{maxo2}). First one
finds $\hat p(x_1|y_3)=55/60$, since this provides the largest possible
value for the joint probability: $\hat p(x_1,y_3)=0.5$. Due to
(\ref{xu2}) this already sets $\hat p(x_1|y_1)=\hat p(x_1|y_2)=0$. Next,
one finds $\hat p(x_2|y_2)=25/30$, since this provides the
second-largest value of the joint probability, $\hat p(x_2|y_2)=0.25$,
also enforcing $\hat p(x_2|y_1)=\hat p(x_2|y_3)=0$. Remaining $p(x|y)$
in (\ref{xu4}) are recovered from normalization. 

It is seen that $\hat p(x|y)$ given in (\ref{xu2}--\ref{xu4}) do have
the maximal number of zeroes (4 for the considered case $n=m=3$) allowed
by (\ref{op}). I.e. the maximizers of $L_{\beta>1}$ are located at
vertices of the convex domain (\ref{op}).

\comment{
\BEA
\label{xu5}
&& (p(y_1), p(y_2), p(y_3))=(0.29, 0.5, 0.21),\\
\label{xu6}
&& p(x_1)=0.6= {\bf \frac{7}{29}}\times 0.29+  {\bf 1}\times 0.5+ {\bf \frac{3}{21}}\times 0.21, \\ 
&& p(x_2)=0.22= {\bf \frac{22}{29}}\times 0.29+ {\bf 0}\times 0.5+ {\bf 0}\times 0.21, \\
&& p(x_3)=0.18= {\bf {0}}\times 0.29+ {\bf 0}\times 0.5+ {\bf \frac{18}{21}}\times 0.21
\label{xu8}
\EEA
}

The second example is dealt with in the same way with $\hat p(x_1|y_1)=4/9$ being the first step, and
$\hat p(x_3|y_2)=\hat p(x_3|y_3)=1$ amount to the last step:
\BEA
\label{xu9}
&& (p(y_1), p(y_2), p(y_3))=(0.9, 0.06, 0.04),\\
\label{xu10}
&& p(x_1)=0.4= {\bf \frac{4}{9}}\times 0.9+  {\bf 0}\times 0.06+ {\bf 0}\times 0.04, \\ 
&& p(x_2)=0.35= {\bf \frac{35}{90}}\times 0.9+ {\bf 0}\times 0.06+ {\bf 0}\times 0.04, \\
&& p(x_3)=0.25= {\bf \frac{15}{90}}\times 0.9+ {\bf 1}\times 0.06+ {\bf 1}\times 0.04.
\label{xu13}
\EEA
The maximizers (but not the
value of $L_\beta$) do not depend on $\beta$ provided that $\beta>1$.
However, we noted that for $\beta\to \infty$ the global maximum of $L_\beta$
are difficult to reach numerically, since they are plagued by many local
maxima. Hence employing moderate values of $\beta$ (e.g. $\beta=2$) can
be beneficial for finding the global maximum numerically. This point can be
illustrated by comparing the global maximizer (\ref{xu10}--\ref{xu13}) with
\BEA
\label{xuu9}
&& (p(y_1), p(y_2), p(y_3))=(0.9, 0.06, 0.04),\\
\label{xuu10}
&& p(x_1)=0.4= {\bf \frac{4}{9}}\times 0.9+  {\bf 0}\times 0.06+ {\bf 0}\times 0.04, \\ 
&& p(x_2)=0.35= {\bf \frac{29}{90}}\times 0.9+ {\bf 1}\times 0.06+ {\bf 0}\times 0.04, \\
&& p(x_3)=0.25= {\bf \frac{21}{90}}\times 0.9+ {\bf 0}\times 0.06+ {\bf 1}\times 0.04.
\label{xuu13}
\EEA
Both (\ref{xu10}--\ref{xu13}) and (\ref{xuu10}--\ref{xuu13}) 
produce the same value for $L_{\beta\to\infty}$, because
\BEA
L_{\beta\to\infty}=\sum_yp(y)\ln\left[\hat p(\widetilde x(y)|y)
\right]+\sum_y p(y)\ln p(y), \qquad
\widetilde x(y)\equiv{\rm argmax}_x[\, p(x|y) \,].
\label{bobo}
\EEA
Indeed, both (\ref{xu10}--\ref{xu13}) and (\ref{xuu10}--\ref{xuu13}) 
have the same values of $\hat p(\widetilde x(y)|y)$.
Even though the global maximum of $L_{\beta>1}$ may be difficult to
reach numerically, we noted that numerically reachable local maxima also have
the same (i.e. maximal) number of zeros. 

\comment{
In this sense the maximation (even local) of $L_{\beta>1}$ leads us to a
sort of Occam's razor, where the most economical explanation is
preferred under given constraints. 
}

\comment{
Let us order: 
\BEA
p(y_1^{\downarrow})\geq p(y_{2}^{\downarrow})\geq ....\geq p(y_m^{\downarrow}), 
\qquad
p(x_1^{\downarrow})\geq p(x_{2}^{\downarrow})\geq ...\geq p(x_n^{\downarrow}).
\EEA

Now the maximization of (\ref{udodo}) proceeds as follows.  First,
$p(x_1^{\downarrow}|y_1^{\downarrow})$ is chosen as large as possible.
The maximization starts from $p(x_1^{\downarrow}|y_1^{\downarrow})$,
because it is multiplied by the largest factor $p(y_1^{\downarrow})$ in
(\ref{udodo}). Next, $p(x_1^{\downarrow}|y_2^{\downarrow})$ is chosen
such that the normalization $p(x_1^{\downarrow})=p(x_1^{\downarrow}|y_1^{\downarrow})p(y_1^{\downarrow})
+p(x_1^{\downarrow}|y_2^{\downarrow})p(y_2^{\downarrow})$ holds with
$p(x_1^{\downarrow}|y_{k>2}^{\downarrow})=0$. Next,
$p(x_2^{\downarrow}|y_1^{\downarrow})$ is chosen as large as possible,
and $p(x_2^{\downarrow}|y_2^{\downarrow})$ is taken such that$p(x_2^{\downarrow})=p(x_2^{\downarrow}|y_1^{\downarrow})p(y_1^{\downarrow})
+p(x_2^{\downarrow}|y_2^{\downarrow})p(y_2^{\downarrow})$ holds. This
procedure is continued till all $p(x|y)$ is fixed. 

\BEA
\label{xxu5}
&& (p(y_1), p(y_2), p(y_3))=(0.29, 0.5, 0.21),\\
\label{xu6}
&& p(x_1)=0.6= {\bf \frac{10}{29}}\times 0.29+  {\bf 1}\times 0.5+ {\bf 0}\times 0.21, \\ 
&& p(x_2)=0.22= {\bf \frac{1}{29}}\times 0.29+ {\bf 0}\times 0.5+ {\bf 1}\times 0.21, \\
&& p(x_3)=0.18= {\bf \frac{18}{29}}\times 0.29+ {\bf 0}\times 0.5+ {\bf 0}\times 0.21
\label{xxu8}
\EEA
}

\end{document}